\documentclass[sigconf]{acmart}
\usepackage{microtype}
\usepackage[american]{babel}
\usepackage{amsfonts}
\usepackage{amsmath}
\usepackage{multirow}
\usepackage{booktabs} 
\usepackage{bm}

\AtBeginDocument{%
  }
\setcopyright{acmlicensed}
\copyrightyear{2025}
\acmYear{2025}
\setcopyright{acmlicensed}\acmConference[KDD '25]{Proceedings of the 31st ACM SIGKDD Conference on Knowledge Discovery and Data Mining V.1}{August 3--7, 2025}{Toronto, ON, Canada}
\acmBooktitle{Proceedings of the 31st ACM SIGKDD Conference on Knowledge Discovery and Data Mining V.1 (KDD '25), August 3--7, 2025, Toronto, ON, Canada}
\acmDOI{10.1145/3690624.3709281}
\acmISBN{979-8-4007-1245-6/25/08}

\begin{document}

\title{Language Representation Favored Zero-Shot Cross-Domain Cognitive Diagnosis}
\author{Shuo Liu}
\authornote{These authors contribute equally to this work.}
\email{shuoliu@stu.ecnu.edu.cn}
\orcid{0000-0001-7970-3187}
\affiliation{%
  \institution{Shanghai Institute of AI Education, and School of Computer Science and Technology\\ East China Normal University}
  \city{Shanghai}
  \country{China}
}

\author{Zihan Zhou}
\authornotemark[1]
\email{zhzhou@stu.ecnu.edu.cn}
\orcid{0009-0009-7784-4846}
\affiliation{%
  \institution{Shanghai Institute of AI Education, and School of Computer Science and Technology\\ East China Normal University}
  \city{Shanghai}
  \country{China}
}

\author{Yuanhao Liu}
\email{51275901044@stu.ecnu.edu.cn}
\orcid{0009-0007-3940-6728}
\affiliation{%
  \institution{Shanghai Institute of AI Education, and School of Computer Science and Technology\\ East China Normal University}
  \city{Shanghai}
  \country{China}
}

\author{Jing Zhang}
\email{jzhang@ed.ecnu.edu.cn}
\orcid{0000-0001-5403-5442}
\affiliation{%
  \institution{Department of Educational Psychology, Faculty of Education\\ East China Normal University}
  \city{Shanghai}
  \country{China}
}

\author{Hong Qian}
\authornote{Hong Qian is the corresponding author.}
\email{hqian@cs.ecnu.edu.cn}
\orcid{0000-0003-2170-5264}
\affiliation{%
  \institution{Shanghai Institute of AI Education, and School of Computer Science and Technology\\ East China Normal University}
  \city{Shanghai}
  \country{China}
}


\begin{abstract}
Cognitive diagnosis aims to infer students' mastery levels based on their historical response logs. However, existing cognitive diagnosis models (CDMs), which rely on ID embeddings, often have to train specific models on specific domains. This limitation may hinder their directly practical application in various target domains, such as different subjects (e.g., Math, English and Physics) or different education platforms (e.g., ASSISTments, Junyi Academy and Khan Academy). To address this issue, this paper proposes the language representation favored zero-shot cross-domain cognitive diagnosis (LRCD). Specifically, LRCD first analyzes the behavior patterns of students, exercises and concepts in different domains, and then describes the profiles of students, exercises and concepts using textual descriptions. Via recent advanced text-embedding modules, these profiles can be transformed to vectors in the unified language space. Moreover, to address the discrepancy between the language space and the cognitive diagnosis space, we propose language-cognitive mappers in LRCD to learn the mapping from the former to the latter. Then, these profiles can be easily and efficiently integrated and trained with existing CDMs. Extensive experiments show that training LRCD on real-world datasets can achieve commendable zero-shot performance across different target domains, and in some cases, it can even achieve competitive performance with some classic CDMs trained on the full response data on target domains. Notably, we surprisingly find that LRCD can also provide interesting insights into the differences between various subjects (such as humanities and sciences) and sources (such as primary and secondary education).
\end{abstract}




\maketitle
\section{Introduction}
\begin{figure}[!t]
\centering
\includegraphics[width=0.75\linewidth]{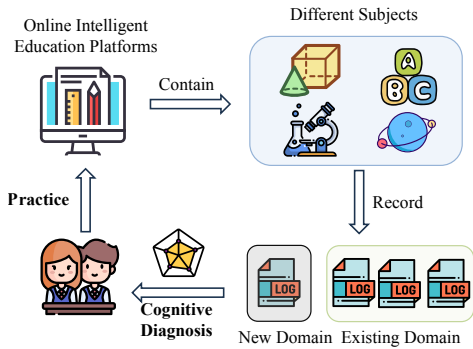}\\
    \caption{An example of zero-shot cross-domain cognitive diagnosis.}
    \label{fig:cd}
\end{figure}

Online intelligent education platforms (OIDP) (e.g., ASSISTments, Junyi Academy and Khan Academy) serve as tools for proactive learning~\cite{Liu2021Towards,Sun2023AKT}, providing personalized practice opportunities that enable students to rapidly enhance their mastery of specific concepts. As shown in Figure~\ref{fig:cd}, cognitive diagnosis (CD)~\cite{Liu2021Towards,LiuZWYL23-fcs}, as a crucial component of these platforms, aims to uncover students' mastery levels (a.k.a., diagnosis results) of specific concepts and the characteristics of exercises based on their historical response logs. The results can support further customized applications, such as exercise recommendation~\cite{Xu2020Course,Gu2024kcr,Yang2023Tkt} and computerized adaptive testing~\cite{Zhuang2022Ncat,zhuang2023BECAT}. 

Over recent years, a diverse array of cognitive diagnosis models (CDMs) has been developed, notably including Item Response Theory (IRT) and the Neural Cognitive Diagnosis Model (NCDM). IRT~\cite{Sympson1978Mirt} employs latent factors to represent mastery levels and utilizes the logistic function as the interaction function (IF) to predict student performance on exercises. In contrast, NCDM~\cite{Wang2020Ncdm}, a pioneering neural-based CDM, replaces the traditional manual IF with multi-layer perceptrons (MLP) and has achieved success in large-scale OIDP. Consequently, neural-based CDMs~\cite{Gao2021Rcd,Ma2022Kscd,Wang2023Kancd, Chen2023Dcd} have rapidly gained prominence. Most existing of them continue the ID-based embedding paradigm, vectorizing students, exercises, and concepts through embeddings and distinguishing them by IDs.

However, as shown in Figure~\ref{fig:cd}, with the increasing diversification in education, students' demands for a variety of subjects are also rising. OIDP now encompasses an increasing number of subjects, ranging from sciences like mathematics and physics to humanities like English and political science. It also caters to a wide range of students, from primary school to university level. However, existing ID-based embedding paradigm forces teachers or researchers to train specific CDMs on specific response logs, causing practical difficulties and inconveniences in applying CDMs across different domains, such as varying subjects and age groups of students. Although some studies~\cite{Gao2023TechCD, Gao2024Z13} have made significant efforts to tackle this task, they still follow the ID-based paradigm or rely on strong assumptions to achieve zero-shot cognitive diagnosis. Nonetheless, these assumptions could be difficult to meet in real educational settings. For instance, TechCD~\cite{Gao2023TechCD} requires that the source domain and the target domain have anchor students, which means that there are some students common to both domains. Similarly, Zero-1-3~\cite{Gao2024Z13} requires early-bird students in the target domain to learn the shared cognitive signals, which can then be transferred to the target domain. 

\textbf{Motivation.} Generally, OIDP already possesses response logs from various domains. When new domain's response logs emerge, it often needs to leverage prior knowledge to quickly and accurately provide diagnostic results for students without retraining models. Therefore, in this paper, we concentrate on a more generalized scenario where one only possesses the response logs from the source domain and lacks any information about the target domain. We designate this critical task as zero-shot cross-domain cognitive diagnosis (ZSCD).

\textbf{Contribution.} To this end, this paper proposes the language representation favored zero-shot cross-domain cognitive diagnosis (LRCD). \textbf{\emph{Our core idea is to articulate the behavior patterns of students, exercises and concepts in response logs from different domains through textual descriptions, referred to as textual cognitive profiles.}} Specifically, LRCD first analyzes the behavior patterns of students, exercises, and concepts in various response data, and then describes the profiles of students, exercises and concepts using textual descriptions. Utilizing recent advanced text-embedding modules (e.g., Llama or OpenAI-3-large), these profiles derived from various domains can be transformed into vectors in the unified language space. Nonetheless, this approach introduces a new challenge, specifically, the inherent discrepancy between the language space and the space of CD. Therefore, we propose language-cognitive mappers in LRCD to learn the mapping from the language space to the cognitive space. Consequently, these profiles can be seamlessly and efficiently integrated and trained with existing CDMs. Notably, we observe that even a simple MLP with two linear layers and ReLU can achieve commendable performance. Extensive experiments demonstrate that training LRCD on most real-world datasets can achieve commendable zero-shot performance on totally different target domains, and in some cases it can even match the performance of classic CDMs trained on the complete target response data. Interestingly, through LRCD's performance across different source domains, we find that data from science subjects tend to be more transferable, demonstrating better performance across various target domains. Additionally, LRCD trained on data from students at higher educational levels exhibit greater transferability when applied to students at lower educational levels in different domains. 

The following sections respectively review the related work, outline the preliminaries, introduce the proposed LRCD, present the empirical analysis, and ultimately conclude the paper. More details of LRCD and experimental setup are provided in the Appendix.

\section{Related Work}
\subsection{Traditional Cognitive Diagnosis}
Cognitive diagnosis has been extensively researched for decades in  educational measurement~\cite{Sympson1978Mirt} or intelligent education~\cite{Liu2021Towards}, aiming to infer students' mastery levels on concepts based on their historical response logs. Item response theory (IRT) and multidimensional IRT (MIRT)~\cite{Sympson1978Mirt} employ latent factors to represent students' mastery levels and use logistic functions as item functions (IF) to predict students' performance on exercises. NCDM~\cite{Wang2020Ncdm}, as a pioneer of neural-based CDM, directly models students' mastery levels on specific concepts with ID-embeddings and employs MLP as IF, which is successful and remarkable. SCD~\cite{DBLP:conf/aaai/ShenQZZ24} introduces the symbolic tree to explicably represent IF to further improve the interpretability. Subsequently, following the ID-embedding paradigm, MLP-based (e.g., CDMFKC~\cite{Li2022CDMFKC}, KSCD~\cite{Ma2022Kscd}, KaNCD~\cite{Wang2023Kancd}), Bayesian network-based~\cite{Li2022Hiercdf}, and GNN-based approaches (e.g., RCD~\cite{Gao2021Rcd}, ORCDF~\cite{QLL2024kdd}) have swiftly achieved even greater success. Recently, a large language model based method called FineCD~\cite{LiQLHZZ25-fcsc} has been proposed to further enhance CD. FineCD incorporates side information such as question statements to realize fine-grained CD. However, due to the absence or scarcity of training data in target domains, CDMs typically focus on training specific models for specific domains, which may hinder their direct application in entirely different target domains (e.g., different subjects or platforms~\cite{oids}).

\subsection{Cross-Domain Cognitive Diagnosis}\label{re:cdcd}
Recently, cross-domain cognitive diagnosis has garnered increasing attention, as the proliferation of subjects on OIDP continues, making it challenging to obtain or access abundant domain-specific data during training. TechCD~\cite{Gao2023TechCD} initially proposed the knowledge concept graph to link concepts across different domains, thus effectively transferring student cognitive signals from source domains to target domains. However, the validity of the hand-crafted knowledge concept graph and the feasibility of directly connecting concepts (e.g., concepts from different subjects) across different domains may significantly influence its cross-domain cognitive diagnosis performance. Zero-1-3~\cite{Gao2024Z13} capitalizes on early bird students in the target domains to learn shared cognitive signals, which can be transferred to the target domain, thereby enriching the cognitive priors for the new domain. However, the necessity of having early bird students in the target domain may not always be feasible in real educational scenarios. More importantly, both TechCD and Zero-1-3 require an overlap of students between the source domain and the target domain, which does not constitute a completely zero-shot cross-domain cognitive diagnosis. In this paper, we focus on a more generalized and challenging task: zero-shot cross-domain cognitive diagnosis, where there is no overlap of information between the source domain and the target domain. We will elaborate on this task in the following sections.

\section{Preliminaries}
Let us consider an intelligent education platform with numerous response logs across \( M \) different domains, which can be formulated as \( R = \{ R_1, R_2, \cdots, R_M \} \). In a specific domain, the response logs consist of a vast number of quadrants, which can be represented as 
\(
R_m = \{(s, e, \{c \mid \mathbf{Q}_{e,c}=1\}, y_{se}) \mid s \in S_m, e \in E_m, c \in C_m, y_{se} \in \{0, 1\}\}.
\)
\( S_m \), \( E_m \), and \( C_m \) denote the sets of students, exercises, and concepts inherent in the domain \( m \), respectively. \(\mathbf{Q}\) represents the relationship between exercises and concepts, where \(\mathbf{Q}_{e,c}=1\) denotes that the exercise \( e \) is related to the concept \( c \). Next, we will present the formal definition of Zero-Shot Cross-Domain Cognitive Diagnosis (ZSCD).

\textbf{Problem Definition of ZSCD.} \label{def:zscd} Suppose there are \( M_o \) source domains' response logs, namely, \( R_o = \{ R_1, R_2, \cdots, R_{M_o} \} \). The goal is to train CDMs on \( R_o \) and infer the mastery level of students in the target domains in a zero-shot manner. Namely, there are no overlapping students, exercises and concepts between the source domain and the target domain.

\begin{figure*}[!t]
\centering
\includegraphics[width=0.99\linewidth]{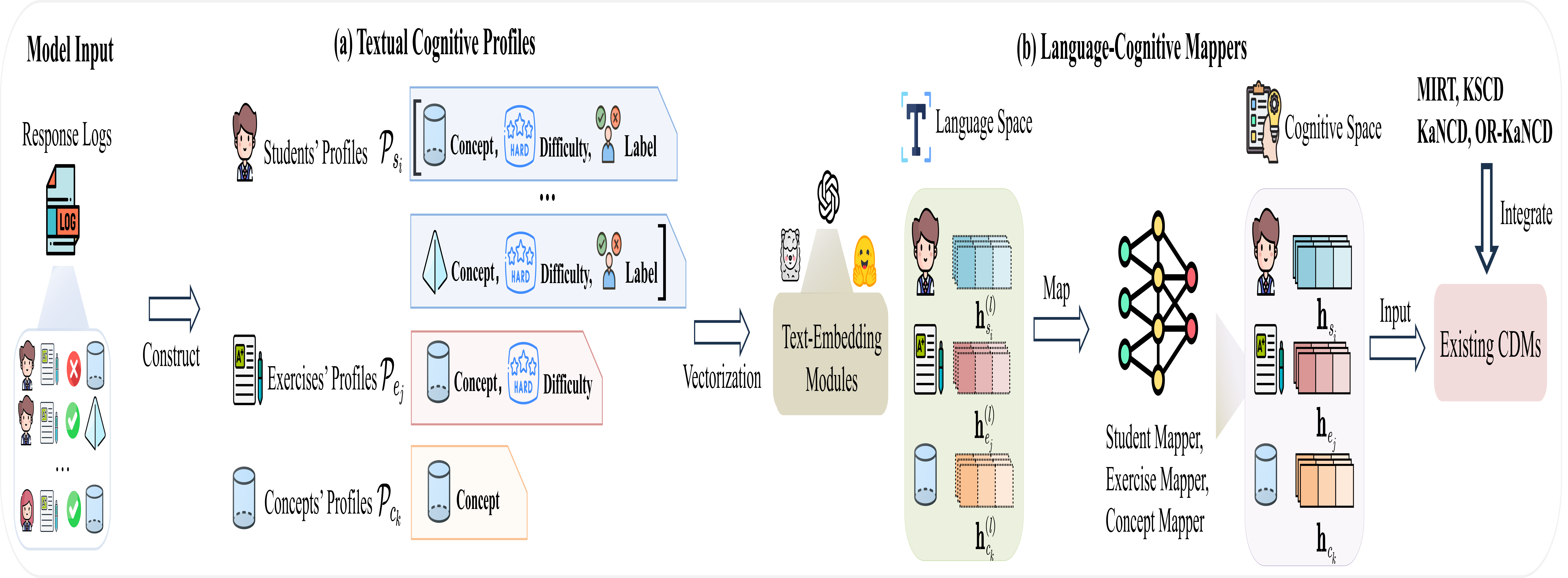}
\caption{An overview of our proposed LRCD framework. The left side (i.e., sub-figure (a)) provides an overview of the proposed Textual Cognitive Profiles. The right side (i.e., sub-figure (b)) shows the proposed Language-Cognitive Mappers.}
\label{fig:framework}
\end{figure*}

\section{Methodology: The Proposed LRCD}
This section introduces the proposed LRCD. It begins by introducing the textual cognitive profiles, specifically how to describe students, exercises, and concepts in different domains. Moreover, leveraging recent advances in text-embedding modules, we can transform these descriptions into vectors within a unified language space. Next, we explore the language-cognitive mapper, a technique designed to map vectors from the language space to the cognitive space. Following this, we explain how these vectors are trained alongside existing diagnostic models. Finally, we discuss how to perform zero-shot inference in a completely new target domain and analyze model complexity. An overview of LRCD is provided in Figure~\ref{fig:framework}.

\subsection{Textual Cognitive Profiles}\label{sec:tcp}
Due to the sensitivity and sparsity of educational data, the input of CD is very simple, consisting only of response logs, where each log contains students' IDs, exercises' IDs, and concept IDs. Recently, researchers have followed the mature paradigm in user modeling by utilizing ID embeddings to represent the features of students, exercises, and concepts~\cite{Wang2020Ncdm, Wang2023Kancd}. Initially, these ID embeddings are randomly initialized. Subsequently, through supervised training, each embedding is updated via back propagation. However, this paradigm results in embeddings trained in different domains being on completely different scales and in different spaces, making them unsuitable for direct application in a different target domain.

To tackle this issue, we need to consider what data can be unified in response logs across different domains. Our core idea is to analyze the behavior patterns of students, exercises, and concepts in response logs and utilize textual descriptions, which we refer to as the \textbf{\emph{textual cognitive profiles}} of response logs. We begin by elucidating the profiles of the concepts.

\textbf{Concepts' Profiles.}
As students use OIDP to assess their mastery of concepts across different domains, it is crucial to effectively articulate the profiles of concepts. 
Different from TechCD~\cite{Gao2023TechCD} who introduces the hand-crafted knowledge concept graph, leveraging concepts as a pivotal bridge to interconnect various domains. Here, we employ the concepts' names as their profiles, as they often reflect their intrinsic interconnections, such as calculus and multivariable calculus. Moreover, these names can be accessed in nearly all OIDPs (e.g.,  ASSISTments). It can be expressed as 
\begin{equation}
\mathcal{P}_{c_k} = \textbf{Name}(c_k) \,,
\end{equation}
where $c_k$ denotes the $k$-th concept and $\mathcal{P}_{c_k}$ is the processed textual profile of $c_k$. $\textbf{Name}$ represents the concept name.

\textbf{Exercises' Profiles.}
Exercises serve as the intermediaries to measure students' mastery of concepts in response logs. By analyzing the response logs, we can infer that it is crucial not only to differentiate exercises based on distinct concepts but also to categorize exercises of the same concept according to their difficulty levels. However, expert labeling of exercise difficulty is a time-consuming and laborious process that often lacks precision. For instance, if a supposedly challenging exercise is administered to a group of high-achieving students, the results may inaccurately suggest that the exercise is not difficult. Consequently, expert labels may fail to accurately reflect the true difficulty of exercises within a specific domain. To address this issue, we utilize the concept names and the average accuracy rates (ACR) of exercises as their profiles. It can formulated as
\begin{equation}
   \text{ACR}_{e_j} = \frac{1}{Z_j} \sum_{i} y_{ij}
\,,\ \mathcal{P}_{e_j} = [\{c_k \mid \mathbf{Q}_{j, k}=1\}, \text{ACR}_{e_j}] \,,
\end{equation}
where $e_j$ denotes the $j$-th exercise. $\text{ACR}_{e_j}$ denotes the average correct rate of $e_j$. $Z_j$ denotes the number of students who have practiced $e_j$. $c_k$ is the related concepts labeled by experts. $\mathcal{P}_{e_j}$ is the processed textual profile of $e_j$. Notably, when calculating $ACR_{e_j}$, we exclusively utilize the available training data to mitigate the risk of information leakage.

\textbf{Students' Profiles.}
 Students' profiles are pivotal to the success of CDMs, as highlighted by recent researchers~\cite{Li2024Idcd, Liu2024Icdm}. We assert that excellent students' profiles should not only consist of their interaction records with exercises but also reflect their mastery of different concepts. However, given that the goal of CD is to infer students' mastery levels across various concepts, this creates a classic ``the chicken or the egg'' dilemma.

Consequently, we integrate the concept name, exercises' average correct rate, and the student score as proxies. Thus, each student-exercise interaction can be considered an integral part of the students' profiles. It can be mathematically expressed as
\begin{equation}
 \mathcal{I}_{ij}=[\{c_k \mid \mathbf{Q}_{j, k}=1\}, \text{ACR}_{e_j}, y_{ij}]\,,\  \mathcal{P}_{s_i} = \{\mathcal{I}_{i0}, \ldots, \mathcal{I}_{ij},\ldots \} \,,
\end{equation}
where $\mathcal{I}_{ij}$ denotes the textual profile of interaction between student $s_i$ and exercise $e_j$, and $y_{ij}$ represents the $i$-th student's score on exercise $e_j$. $\mathcal{P}_{s_i}$ is the processed textual profile of $s_i$. Notably, unlike other profiles, $\mathcal{P}_{s_i}$ is a set comprising the processed textual profiles of interactions by student $s_i$.

\subsection{Language-Cognitive Mappers}\label{sec:lcm}
\textbf{Vectorization.} Obviously, through the aforementioned methods, we can analyze each response log and convert them into textual cognitive profiles. Although these textual descriptions cannot be directly used in existing CDMs, we can leverage advanced text-embedding modules (e.g., OpenAI-3-large) to transform these profiles into vectors in the language space which can be formulated as
\begin{equation}
\mathbf{h}_{s_i}^{(l)}=\sum_{j}\frac{\textbf{TEM}(\mathcal{I}_{ij})}{|\mathcal{P}_{s_i}|},\, 
   \mathbf{h}_{e_j}^{(l)}=\textbf{TEM}(\mathcal{P}_{e_j}),\,
   \mathbf{h}_{c_k}^{(l)}=\textbf{TEM}(\mathcal{P}_{c_k})\,,
\end{equation}
where $\mathbf{h}_{s_i}^{(l)},\, \mathbf{h}_{e_j}^{(l)},\, \mathbf{h}_{c_k}^{(l)} \in \mathbb{R}^{1 \times d_l}$ represent the vectors of student $s_i$, exercise $e_j$ and concept $c_k$ in the language space, respectively. $d_l$ is the dimension of the text embedding (e.g., 3072 in OpenAI-3-large). $\textbf{TEM}$ denotes the text-embedding module (e.g., OpenAI-3-large) which can be seen as hyperparameters in LRCD, we give further experiments in Section. The representation of $s_i$ can be considered as the mean pooling of the text embeddings of each of his or her interactions. Notably, we exclusively utilize the available training interactions to mitigate the risk of information leakage. Detailed information can be found in Section~\ref{sec:exp}.

\textbf{Mappers.} Leveraging the aforementioned text cognitive profiles and advanced text-embedding modules, we can harmonize data from diverse domains into a unified language space. This enables our model to diagnose students' abilities without the necessity of training on target-domain data. However, the language space is substantially disparate from the space of CD, as the former is trained on extensive corpora (e.g., Common Crawl, Wikipedia, and BooksCorpus) that are entirely unrelated to education data. Therefore, we propose language-cognitive mappers to learn the projection of the language space to the cognitive space which can be formulated as
\begin{equation}
\mathbf{h}_{s_i}=\mathcal{F}_s(\mathbf{h}_{s_i}^{(l)};\bm{\theta}_s)\,,\ 
   \mathbf{h}_{e_j}=\mathcal{F}_e(\mathbf{h}_{e_j}^{(l)};\bm{\theta}_e)\,,\ 
\mathbf{h}_{c_k}=\mathcal{F}_c(\mathbf{h}_{c_k}^{(l)};\bm{\theta}_c) \,,
\end{equation}
where $\mathbf{h}_{s_i}, \mathbf{h}_{e_j}, \mathbf{h}_{c_k} \in \mathbb{R}^{1 \times d}$ represent the vectors of student $s_i$, exercise $e_j$ and concept $c_k$ in the space of CD. $d$ is the dimension of space of CD (e.g. 32 in KaNCD~\cite{Wang2023Kancd}). $\mathcal{F}_s$, $\mathcal{F}_e$ and $\mathcal{F}_c$ indicate the student mapper, exercise mapper and concept mapper where $\bm{\theta}_s,\bm{\theta}_e,\bm{\theta}_c$ are model parameters. Concretely, each mapper follows the identical model architecture. Details can be found in Section~\ref{sec:exp}.

\subsection{Training and Inference}
\textbf{Model-Agnostic.} Given that the primary focus of LRCD is to handle ZSCD tasks, we do not design an additional CDM. Instead, we incorporated recent advanced CDMs into LRCD to illustrate its efficacy and adaptability. It can be expressed as
\begin{equation}
   \hat{y}_{ij}=\mathcal{M}_{\text{CD}}(\mathbf{h}_{s_i},\mathbf{h}_{e_j}, \mathbf{h}_{c}; \bm{\theta}_{\text{CD}}) \,,
\end{equation}
where $\hat{y}_{ij}$ is the score prediction of student $s_i$ practice exercise $e_j$. $\mathcal{M}_{\text{CD}}$ denotes the integrated CDM (e.g., KSCD~\cite{Ma2022Kscd}, KaNCD~\cite{Wang2023Kancd}, ORCDF~\cite{QLL2024kdd}). \(\mathbf{h}_{s_i}, \mathbf{h}_{e_j}, \mathbf{h}_{c}\) indicate the representations of \(s_i\), \(e_j\), and concepts, which are the outputs of LRCD. $\bm{\theta}_{\text{CD}}$ represents the parameters of the integrated CDM.

\textbf{Multi-Domain Training.} Consistent with previous papers~\cite{Wang2020Ncdm}, we employ supervised learning to recover the response logs while simultaneously estimating the whole model parameters (i.e., $\bm{\theta}_{s},\bm{\theta}_{e},\bm{\theta}_{c},\\ \bm{\theta}_{\text{CD}}$). Specifically, we compute the loss between the model's predictions and the actual response scores within a mini-batch and utilize binary cross-entropy (BCE) as the loss function. Notably, LRCD can be trained on multi-domain data, suppose that there are $M_o$ source domains as introduced in Section~\ref{def:zscd}, it can be formulated as follows
\begin{equation}
    \begin{aligned}
    &\mathcal{L}_{R_m} = - \sum_{(s,e,c, y_{se}) \in R_m} \left[ y_{se} \log \hat{y}_{se} + (1 - y_{se}) \log (1 - \hat{y}_{se}) \right]\,, \\
    &\mathcal{L} = \sum_m^{|R_{o}|} w_m \mathcal{L}_{R_m}\,.
    \end{aligned}
\end{equation}
$\mathcal{L}_{R_m}$ denotes the BCE loss of domain $R_m$. $w_m$ is the weight of loss in domain $m$. In implementation, we directly choose \( w \) as a constant value, namely $\frac{1}{|R_{o}|}$. For more sophisticated selections, we defer to future work.

\textbf{Zero-Shot Inference in Target Domains.} Once the training of LRCD reaches convergence, we can undertake zero-shot inference in any target domain. We first \textbf{frozen the parameters in LRCD}. Then, we tackle the response logs in the train data of target domains into textual cognitive profiles and vectorized them with identical TEM. Benefiting from the proposed mappers, we can rapidly acquire the representations of students, exercises, and concepts in a completely distinct target domain. Ultimately, through the integrated CDM, we can predict the student score on certain exercises or infer students' mastery levels and exercises' difficulty levels.

\subsection{Discussion}\label{sec:discuss}
In this section, we engage in some pivotal discussions about the proposed LRCD, namely time complexity, scalability, and distinction from previous CDMs.

\textbf{Time Complexity Analysis.} Analyzing the time complexity of the CDM is crucial, as students often wish to quickly receive their diagnostic results after completing exercises. Since we do not design a tailored CDM and instead integrate existing CDMs, we will focus our discussion on the time complexity of obtaining text embeddings and the proposed language-cognitive mappers. Favored by the fast API provided by OpenAI~\cite{Radford2019Language}, we can obtain all representations of a normal domain within 10 minutes. \textbf{\emph{More importantly, we only need to run this process once and store the results locally. Therefore, the runtime for this aspect is practically negligible.}} The time complexity of the proposed language mappers is approximately \(O(d_ld)\) which depends on the chosen text-embedding module and integrated CDMs. We will provide further details to illustrate the actual training time and inference time for LRCD in Appendix~\ref{appd:time:train} and Appendix~\ref{appd:time:infer}. Notably, LRCD can directly infer students' mastery levels in a new domain with 1,500 students and 50,000 response logs in just \textbf{\emph{0.1 seconds}}, making it \textbf{\emph{406 times}} faster than model retraining.

\textbf{Scalability.} As we do not design computationally intensive modules (e.g., no need for language model inference~\cite{SDPO}), LRCD is both efficient and easy to implement, allowing for the seamless integration of all existing CDMs. \textbf{\emph{Thus, LRCD can handle thousands of students, exercises and concepts with millions of response logs}}, and we will verify this capability in experiments.

\textbf{Distinction with Zero-1-3~\cite{Gao2024Z13}.} Herein, we primarily focus on discussing the differences between LRCD and Zero-1-3 at the model level, while the differences in tasks are mentioned in Section~\ref{re:cdcd}. Compared with Zero-1-3, which uses exercise content~\cite{liu2023simplekt, Liu2021Ekt} as exercise features and employs pre-trained CDMs to initialize student representations, we argue that exercise descriptions are highly variable, as questions on the same concept can take numerous forms. Therefore, as proposed in our textual cognitive profiles, we describe students, exercises, and concepts through the analysis of their behavior patterns in the response logs. Therefore, we do not need the pre-trained CDMs, and we can place all features into a unified language space using text embedding, thereby successfully achieving ZSCD.

\section{Experiments}\label{sec:exp}
This section first introduces three real-world datasets and evaluation metrics. Subsequently, through comprehensive experiments, we endeavor to substantiate the superiority of LRCD's zero-shot performance across various target domains (e.g., different subjects and different platforms). To ensure the reliability and reproducibility of our experiments, they are independently repeated ten times with different seeds, and our code is available at \url{https://github.com/ECNU-ILOG/LRCD}.

\subsection{Experimental Settings} 
\textbf{Datasets Description.} We conduct our experiments on three real-world datasets: SLP~\cite{Lu2021slp}, EDM~\cite{edm-cup-2023}, and MOOC~\cite{Yu2023MOOCRadar}. To ensure that each student has a sufficient number of diagnostic records and meets the requirements of the baseline as well as equipment needs, we exclude students with fewer than 5, 20 and 50 responses, respectively. Moreover, following the paradigm established by~\cite{Wang2020Ncdm, Li2022Hiercdf}, we restrict to the first attempt for each exercise to ensure the stability of students' mastery levels. This approach ensures that the data reflect the students' initial understanding, thereby maintaining the static nature of their mastery levels. Importantly, we have uploaded all the processed data to the aforementioned repository. Table~\ref{tab:dataset_1} provides detailed statistics of those datasets, where ``Average Correct Rate'' indicates the average accuracy of students on exercises, and ``Q Density'' represents the average number of concepts per exercise. Moreover, the entry for SLP represents the aggregate values across all subjects (e.g., Math, Physics). Due to space constraints, we provide the detailed values for each subject in Table~\ref{tab:dataset_slp}.

\begin{table}[!htbp]
  \centering
  \caption{Statistics of the three real-world datasets.}
 \resizebox{0.80\linewidth}{!}{\begin{tabular}{l|ccc}
    \toprule
    Datasets & SLP & MOOC & EDM \\
    \midrule
    \#Students & 7,663 & 6,077  & 6,061 \\
    \#Exercises & 4,873 & 2,366  & 1,534 \\
    \#Concepts & 179 & 2,611  & 323 \\
    \#Response Logs & 299,391 & 769,541 & 198,058 \\
    Average Correct Rate & 0.484 & 0.829 & 0.601 \\
    Q Density & 1.000 & 2.233 & 1.000 \\
    \bottomrule
\end{tabular}}
  \label{tab:dataset_1}%
\end{table}

\textbf{Evaluation and Settings.}
To evaluate the efficacy of LRCD, we adopt the methodology of previous research by assessing the predictive accuracy of student performance, given that the true mastery levels of students are inherently unobservable in real-world contexts. Consistent with previous studies~\cite{Wang2020Ncdm, Wang2023Kancd}, we validate the accuracy of the diagnostic outcomes produced by CDMs by predicting students' performance on assessments. We employ both performance prediction and interpretability metrics~\cite{Wang2020Ncdm, Sun2024interpretable} to measure effectiveness. Concretely, for the score prediction metric, given that the task is a binary classification problem~\cite{Wang2020Ncdm}, we employ the Area Under the Curve (AUC) as our evaluation metric. For the interpretability metric, in alignment with previous methodologies~\cite{Li2022Hiercdf}, we utilize the degree of agreement (DOA) to assess the interpretability of the inferred mastery levels. For a more detailed explanation of the DOA, please refer to Appendix~\ref{appd:exp:doa}. 

For evaluation, we follow the methodology of previous work~\cite{Wang2023Kancd, Liu2024Icdm}, dividing the response logs of each domain similarly. Specifically, the response logs of each student are partitioned into three parts: 70\% for training, 20\% for validation, and 10\% for testing. During the training phase, we utilize the training data from the source domains to train the model and determine the best hyperparameters using the validation data. During the inference phase, we use the training data in the target domain to infer the students' mastery levels and evaluate the results on the test data. Notably, the training data in the target domain are not available during the training phase, thus preventing information leakage.

Then, to evaluate the performance of LRCD in zero-shot cross-domain cognitive diagnosis, we conduct comparisons under the following four experimental settings. We solely compare Zero-1-3 in the third setting as it requires overlapping students between the source domain and the target domain.

$\bullet$ \textbf{Subject-Level Zero-Shot Student Score Prediction.} In this setting, the source domain and the target domain encompass distinct academic subjects, yet both are derived from the same platform, SLP. For instance, the source domains include Math and Physics, while the target domain is English. 

$\bullet$ \textbf{Platform-Level Zero-Shot Student Score Prediction.}
In this setting, the source domain and the target domain encompass distinct intelligent education platforms, yet both include the same subject, Math. For instance, the source domain is SLP, while the target domain is EDM.

$\bullet$ \textbf{Student Score Prediction with Overlap Students.} In this setting, the source domain and the target domain encompass distinct academic subjects, yet both are derived from the same platform, SLP. Different from the first setting, here we consider scenarios where there is an overlap of students between the source domain and the target domain. 

$\bullet$ \textbf{Standard Setting.} In this setting, the source domain and the target domain are identical. For brevity, we only consider three domains (i.e., SLP-Math,  MOOC, EDM) within the same subject, Math. Due to space limitations, the results are presented in Appendix~\ref{appd:exp:standard}.

\textbf{Baseline Description.} Here, we will introduce our baselines in detail as follows:

$\bullet$ Oracle: It is trained using training data from the target domains, employing integrated traditional CDMs which stands for the upper bound of the performance.

$\bullet$ NCDM~\cite{Wang2020Ncdm}: This is a recent classic neural-based CDM. In this study, we train it using the training data from the target domains to compare it with LRCD's zero-shot performance.

$\bullet$ Random: The embeddings of integrated traditional CDMs in the target domain are randomly initialized with values between 0 and 1 which stands for the lower bound of the performance.

$\bullet$ TechCD ~\cite{Zhuo2022Tiger, Gao2023TechCD}: It uses a pedagogical knowledge concept graph as a mediator to connect students in the source domain with those in the target domain, thus effectively transferring student cognitive signals from source domains to target domains. Following~\cite{Gao2024Z13}, we utilize the statistical method proposed in~\cite{Gao2021Rcd} to construct the graph.

$\bullet$ GCN-based~\cite{Zhuo2022Tiger, Gao2023TechCD}: Derived from TechCD, the original TechCD obtains more transferable representations by pruning the outputs of some lower GCN layers. The GCN-based method is an ablation of TechCD that directly uses only the embeddings from the final layer of nodes.

$\bullet$ NLP-based~\cite{Zhuo2022Tiger, Gao2023TechCD}: Derived from TechCD, it utilizes learnable embeddings for students and encodes the texts of exercises with OpenAI-3-large~\cite{Radford2019Language} to represent them.

$\bullet$ Zero-1-3~\cite{Gao2024Z13}: It employs dual regularizers to decouple student embeddings into domain-shared and domain-specific parts. Additionally, it devises a strategy to generate simulated practice logs for new students in the target domain by analyzing the behavioral patterns of early-bird students in the same domain.

\textbf{Implementation Details.} All parameters are initialized using Xavier initialization and optimized with Adam~\cite{Kingma2015Adam} optimizer. We set $d$ as 64 which is the dimension of vectors utilized in integrated CDMs, and set the batch size to 256. For a fair comparison between LRCD and all baselines, we utilize OpenAI-3-large~\cite{Radford2019Language} as the default text-embedding module and OR-KaNCD, proposed in ORCDF~\cite{Wang2023Kancd}, as the default integrated CDM. The dimensions of the MLP for all methods are consistent, 512 and 256, respectively. We provide a detailed hyperparameter analysis regarding the selection of text-embedding modules and integrated CDMs in Section~\ref{sec:hyper}. The learning rate is chosen from $\{1e^{-5}, 5e^{-5}, 1e^{-4}, 2.5e^{-4}, 5e^{-4}, 2e^{-3} \}$. All experiments are run on a Linux server with two 3.00GHz Intel Xeon Gold 6354 CPUs and two RTX3090 GPUs. All models are implemented by PyTorch. Details about baselines can be found in Appendix~\ref{appd:exp:baselines}.

\begin{table*}[!htbp]
  \centering
\caption{Overall student score prediction performance in subject-level zero-shot cross-domain cognitive diagnosis. Within each method, the highest mean value is highlighted in bold, and the runner-up is underlined. The value following ``$\pm$'' represents the standard deviation of the model's performance. If the mean value significantly differs from the runner-up, passing a $t$-test with a significance level of 0.01, it is marked with ``*''. We use the first letter of each subject to represent it. For example, PB-M signifies that the source domains are Physics and Biology, while the target domain is Math.}
     \resizebox{0.78\linewidth}{!}{
\begin{tabular}{c|c|ccccc|cc}
    \toprule
    Datasets & Metrics & Random & GCN   & NLP   & TechCD & \textbf{LRCD} & NCDM  & Oracle \\
    \midrule
    \multirow{2}[2]{*}{PB-M} & AUC (\%) & 50.03$\pm$0.75 & 51.57$\pm$0.03 & 48.86$\pm$0.03 & \underline{52.66$\pm$0.03} & \textbf{80.23}$^*$$\pm$0.15 & 81.18$\pm$0.09 & 81.74$\pm$0.10 \\
          & DOA (\%) & \underline{51.32$\pm$0.00} & 50.83$\pm$2.11 & 49.41$\pm$1.03 & 50.59$\pm$2.33 & \textbf{77.08}$^*$$\pm$0.06 & 81.68$\pm$0.02 & 81.31$\pm$0.08 \\
    \midrule
    \multirow{2}[2]{*}{HGE-M} & AUC (\%) & 49.24$\pm$2.31 & 50.82$\pm$0.01 & 50.00$\pm$0.00 & \underline{52.53$\pm$0.04} & \textbf{79.54}$^*$$\pm$0.16 & 81.18$\pm$0.09 & 81.74$\pm$0.10 \\
          & DOA (\%) & 50.66$\pm$0.81 & 50.58$\pm$3.78 & 50.48$\pm$1.35 & \underline{53.82$\pm$1.53} & \textbf{76.70}$^*$$\pm$0.09 & 81.68$\pm$0.02 & 81.31$\pm$0.08 \\
    \midrule
    \multirow{2}[2]{*}{PB-C} & AUC (\%) & 50.41$\pm$1.72 & 49.89$\pm$0.02 & 49.96$\pm$0.01 & \underline{51.68$\pm$0.02} & \textbf{83.32}$^*$$\pm$0.23 & 83.47$\pm$0.07 & 84.30$\pm$0.04 \\
          & DOA (\%) & 49.51$\pm$0.43 & 45.09$\pm$2.78 & 50.14$\pm$1.65 & \underline{52.36$\pm$4.38} & \textbf{75.53}$^*$$\pm$0.07 & 81.97$\pm$0.06 & 80.39$\pm$1.17 \\
    \midrule
    \multirow{2}[2]{*}{HGE-C} & AUC (\%) & 50.12$\pm$1.79 & 50.46$\pm$0.02 & \underline{50.84$\pm$0.02} & 50.38$\pm$0.01 & \textbf{83.10}$^*$$\pm$0.22 & 83.47$\pm$0.07 & 84.30$\pm$0.04 \\
          & DOA (\%) & 49.64$\pm$0.66 & 44.97$\pm$2.96 & \underline{50.44$\pm$0.87} & 48.61$\pm$4.81 & \textbf{75.12}$^*$$\pm$0.18 & 81.97$\pm$0.06 & 80.39$\pm$1.17 \\
    \bottomrule
    \end{tabular}%

    }
  \label{tab:exp_subject}%
\end{table*}%

\begin{table*}[!htbp]
  \centering
  \caption{Overall student score prediction performance in platform-level zero-shot cross-domain cognitive diagnosis. Details are the same as Table~\ref{tab:exp_subject}.}
   \resizebox{0.78\linewidth}{!}{ \begin{tabular}{c|c|ccccc|cc}
    \toprule
    Datasets & Metrics & Random & GCN   & NLP   & TechCD & \textbf{LRCD}  & NCDM  & Oracle \\
    \midrule
    \multirow{2}[2]{*}{EDM-MATH} & AUC (\%) & 49.09$\pm$0.65 & 51.44$\pm$0.03 & 51.30$\pm$0.03 & \underline{51.70$\pm$0.03} & \textbf{79.76}$^*$$\pm$0.33 & 81.18$\pm$0.09 & 81.74$\pm$0.10 \\
          & DOA (\%) & 50.61$\pm$1.51 & \underline{54.90$\pm$6.09} & 50.03$\pm$0.68 & 52.88$\pm$2.66 & \textbf{76.92}$^*$$\pm$0.33 & 81.68$\pm$0.02 & 81.31$\pm$0.08 \\
    \midrule
    \multirow{2}[2]{*}{MATH-EDM} & AUC (\%) & 49.37$\pm$0.94 & \underline{51.09$\pm$0.01} & 50.35$\pm$0.03 & 50.17$\pm$0.03 & \textbf{79.41}$^*$$\pm$0.15 & 80.38$\pm$0.07 & 83.48$\pm$0.04 \\
          & DOA (\%) & 49.65$\pm$0.27 & 50.53$\pm$0.89 & 49.59$\pm$0.89 & \underline{50.57$\pm$3.15} & \textbf{77.17}$^*$$\pm$0.14 & 85.42$\pm$0.02 & 79.64$\pm$0.88 \\
    \midrule
    \multirow{2}[2]{*}{MOOC-MATH} & AUC (\%) & \underline{51.39$\pm$3.67} & 50.17$\pm$0.03 & 48.30$\pm$0.03 & 49.99$\pm$0.00 & \textbf{77.06}$^*$$\pm$0.34 & 81.19$\pm$0.09 & 81.74$\pm$0.10 \\
          & DOA (\%) & 50.05$\pm$0.94 & 47.25$\pm$0.00 & 50.21$\pm$1.24 & \underline{50.82$\pm$1.52} & \textbf{75.25}$^*$$\pm$0.46 & 81.68$\pm$0.02 & 81.31$\pm$0.08 \\
    \midrule
    \multirow{2}[2]{*}{MATH-MOOC} & AUC (\%) & 48.92$\pm$8.06 & 50.13$\pm$0.02 & \underline{51.26$\pm$0.01} & 50.15$\pm$0.01 & \textbf{81.57}$^*$$\pm$0.97 & 81.97$\pm$0.06 & 89.63$\pm$0.05 \\
          & DOA (\%) & 50.14$\pm$0.60 & \underline{50.37$\pm$0.06} & 49.93$\pm$0.22 & 50.13$\pm$3.37 & \textbf{75.87}$^*$$\pm$0.85 & 83.52$\pm$0.05 & 80.05$\pm$0.31 \\
    \bottomrule
    \end{tabular}%
    }
  \label{tab:exp_platform}%
\end{table*}%

\subsection{Subject Level Zero-Shot Student Score Prediction}
In this subsection, we will compare the performance of our proposed LRCD with other baselines for Student Score Prediction at the subject level. Specifically, the source domains and target domains are derived from datasets pertaining to different academic subjects (e.g., Math, English). Here, for brevity, we denote each subject by its initial letter, for instance, M for Math. Consequently, PB-M signifies that the source domains are Physics and Biology, while the target domain is Math. Notably, we categorize different subjects based on their attributes into Humanities (History, Geography, English, Chinese) and Sciences (Physics, Biology) to analyze the influence between subjects.

\textbf{Results.} As shown in Table~\ref{tab:exp_subject}, we can obtain two key observations. LRCD significantly outperforms other baselines, \textbf{\emph{achieving at least 97.30\% of the oracle performance and demonstrating competitive results with NCDM in certain scenarios}}. This indicates that LRCD is highly effective in subject-level zero-shot cross-domain cognitive diagnosis. TechCD generally performs better than other baselines but still exhibits suboptimal performance. This may be due to the hand-crafted knowledge concept graph, which links completely unrelated concepts across different subjects, potentially hindering its effectiveness in cross-domain scenarios. Moreover, we can find some interesting discoveries: the cross-domain performance is better when the source domain is a science subject compared to when it is a humanities subject, regardless of whether the target domain is a humanities or science subject. This may be due to the greater inherent differences in concepts within humanities subjects, whereas science subjects tend to have relatively smaller disparities. Science data may be more universally applicable in ZSCD. Of course, we cannot rule out the influence of the data size within each subject. Nonetheless, we believe that this warrants further investigation in future work.


\begin{table}[htbp]
  \centering
  \caption{Ablation Study of LRCD. Details are the same as Table~\ref{tab:exp_subject}.}
 \resizebox{0.80\linewidth}{!}{\begin{tabular}{c|c|c|c|c}
    \toprule
    Datasets & Metrics & LRCD-w/o-TCP & LRCD-w/o-LCM & \textbf{LRCD} \\
    \midrule
    \multirow{2}[0]{*}{PB-M} & AUC (\%) & 75.74 & 61.32 & $\textbf{80.30}^*$ \\
          & DOA (\%) & 71.44 & 71.20 & $\textbf{76.75}^*$ \\
    \midrule
    \multirow{2}[1]{*}{PB-C} & AUC (\%) & 80.67 & 79.10 & $\textbf{83.78}^*$ \\
          & DOA (\%) & 72.94 & 72.81 & $\textbf{76.25}^*$ \\
    \midrule
    \multirow{2}[2]{*}{EHG-C} & AUC (\%) & 80.38 & 77.09 & $\textbf{83.17}^*$ \\
          & DOA (\%) & 70.86 & 65.68 & $\textbf{75.72}^*$ \\
    \bottomrule
\end{tabular}}
  \label{tab:ab}%
\end{table}%

\subsection{Platform Level Zero-Shot Student Score Prediction}
In this subsection, we will compare the performance of our proposed LRCD with other baselines for Student Score Prediction at the platform level. Specifically, the source domains and target domains are derived from datasets pertaining to different education platforms (e.g., MOOC, SLP). For brevity, we use MATH to represent Math from SLP.

\textbf{Results.} As shown in Table~\ref{tab:exp_platform}, LRCD significantly outperforms other baselines, \textbf{\emph{achieving at least 94.91\% of the oracle performance and demonstrating competitive results with NCDM in certain scenarios}}. This indicates that LRCD is highly effective in platform-level zero-shot cross-domain cognitive diagnosis. At the same time, we can discover from the experimental results that when the source domain or target domain is EDM, the model performs better. This may be due to the higher similarity between the exercise on the MATH and EDM platforms. The exercises on these two platforms are entirely within the field of mathematics and are similar in difficulty, leaning towards simpler elementary or middle school math problems. This makes students' problem-solving patterns more similar, which allows the model to perform better on EDM using the data patterns extracted from MATH and vise versa. On the other hand, the MOOC includes not only math problems, but also issues from other fields such as physics and economics. Moreover, the MOOC platform is aimed at older students, typically college students, which means the exercise tends to be more challenging and involve more complex mathematical problems. The significant difference in difficulty between the math problems in MATH and MOOC makes it harder for the model to learn mutually between these two platforms. 

\subsection{Student Score Prediction with Overlap Students}

This subsection compares the performance of the proposed LRCD with other baselines for Student Score Prediction on the same platform SLP. Unlike previous experiments, this time we have constructed a new dataset from SLP, referred to as $\text{SLP}^\dag$, where the source domain and the target domain have overlapping students, consistent with the requirements of Zero-1-3 and TechCD. It comprises 107 students, 2,239 exercises, 93 concepts, and 17,955 response logs. The source domains encompass Biology and Physics, whereas the target domain is Math. Detailed implementations of the baselines are provided in Appendix~\ref{appd:exp:baselines}.

\begin{figure}[!t]
\centering
\begin{minipage}{0.44\linewidth}\centering
    \includegraphics[width=\textwidth]{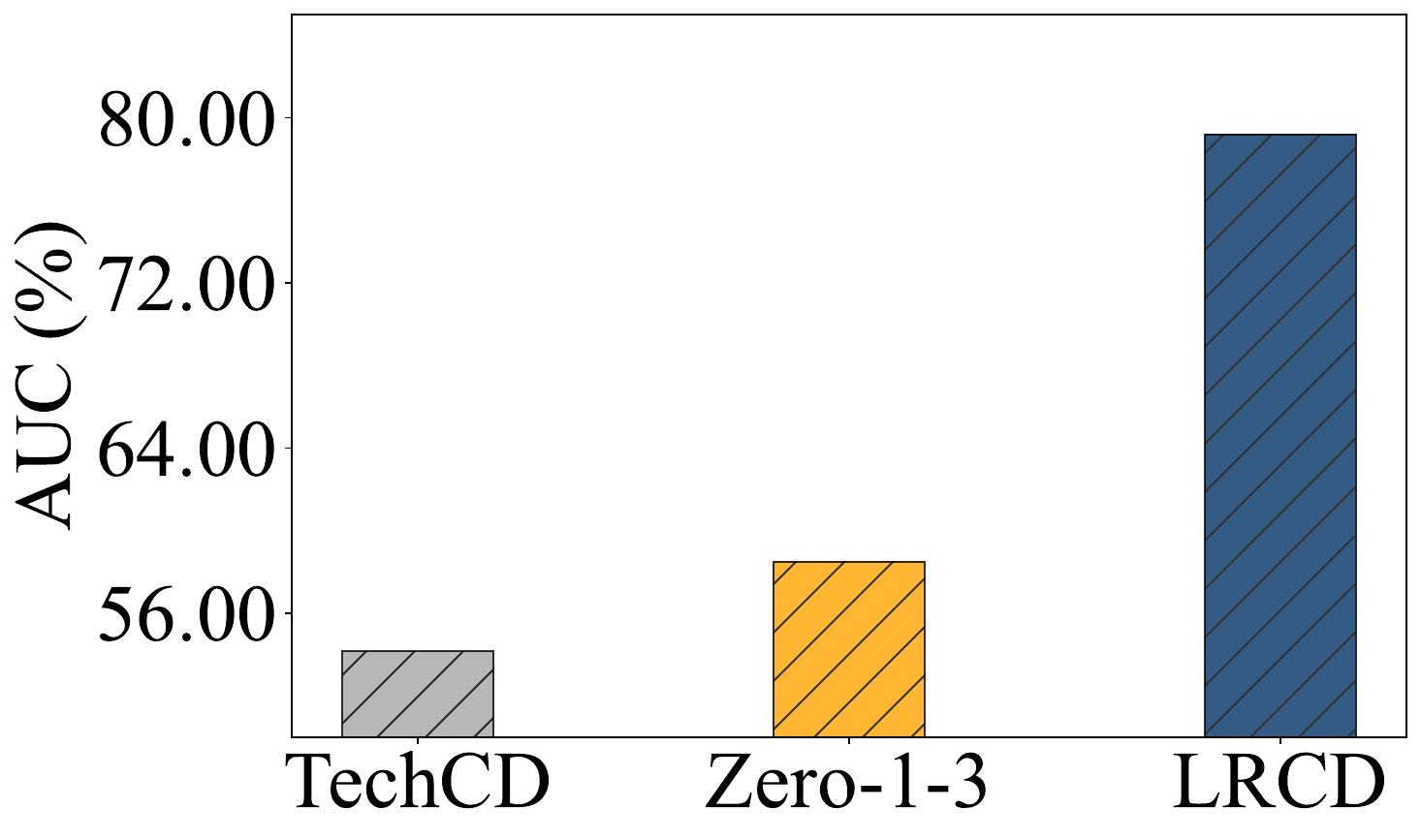}\\
\end{minipage}
\begin{minipage}{0.44\linewidth}\centering
    \includegraphics[width=\textwidth]{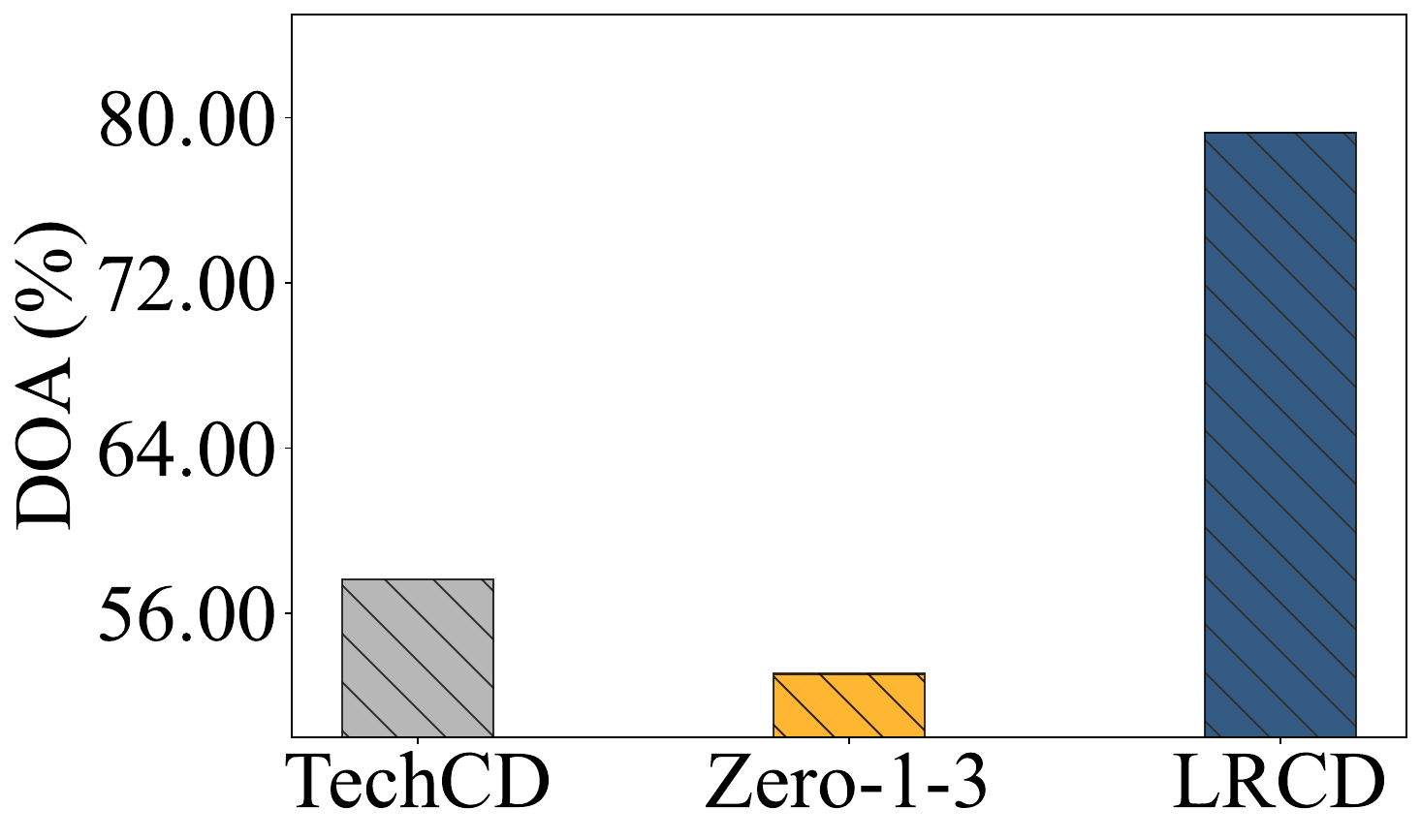}\\
\end{minipage}
\caption{Overall student score prediction performance in overlap students setting on $\text{SLP}^\dag.$}
\label{fig:over1}
\end{figure}

\textbf{Results.} As shown in Figure~\ref{fig:over1}, both TechCD and Zero-1-3 benefit from the overlapping student setting, consistent with their original claims. Their performance has significantly improved compared with previous results. Moreover, LRCD significantly outperforms TechCD and Zero-1-3, achieving an improvement of nearly \textbf{35\%}. This further validates the effectiveness of LRCD in settings with overlapping students.

\subsection{Ablation Study}
To validate the efficacy of the Text Cognitive Profiles (TCP) and Language-Cognitive Mappers (LCM) in LRCD, we conduct an ablation study. We present two ablated versions of LRCD: LRCD-w/o-TCP and LRCD-w/o-LCM. Specifically, the former replaces the TCP with a random vector sampled from a standard normal distribution. The latter omits the use of Mappers and directly employs the obtained text embeddings in the language space. Since the dimension of OpenAI-3-large is 3072, which is excessively large, omitting the proposed LCM would result in high GPU demands. Therefore, we use Bert embeddings as a substitute. 
Unfortunately, LRCD-w/o-LCM still exceeds memory limits in some cases, which also underscores the importance of the proposed LCM. Therefore, we will only report the results for the successful instances.

\textbf{Results.} As illustrated in Table~\ref{tab:ab}, it is evident that LRCD markedly outperforms both ablated versions, substantiating the synergistic effect of integrating both modules. Furthermore, each ablated version of LRCD surpasses the best-performing baseline in Table 1, corroborating the efficacy of each individual module. Surprisingly, we find that LRCD-w/o-TCP also achieves commendable performance, demonstrating the effectiveness of the LRCD framework. This underscores the importance of representing students, exercises, and concepts within the unified space for the ZSCD task. The subpar performance of LRCD-w/o-LCM underscores the considerable disparity between the language space and the cognitive space. This highlights the inadequacy of directly utilizing representations from the language space and emphasizes the critical importance of establishing a mapping between the two spaces.

\subsection{Scaling Up in Datasets}\label{app:exp:sacle}
This subsection aims to investigate whether expanding the scope of the source domain can enhance the performance of LRCD in the target domain. In Table~\ref{appd:fig:scale1}, when our target domain, Math, is included in the training set, we can see that as we continue to add questions from different related domains to the dataset, the model's performance improves steadily. This suggests that adding more related domain questions to the training set can enhance the predictive performance of the target domain in this situation. In Table~\ref{appd:fig:scale2}, our target domain, Chinese, is not included in the training set. When we add questions from related domains to the training set, the model's performance shows some fluctuation. This may be because the model is relatively sensitive to the types of source domain in the training set. Adding data that are not closely related to the target domain might affect performance. However, such additions can also enhance the model's generalization ability to some extent, improving the model's robustness and leading to an overall upward trend in performance.

\begin{table}[!t]
  \centering
  \caption{Scaling up in training datasets to predict SLP-Math.}
    \resizebox{0.75\linewidth}{!}{\begin{tabular}{c|c|ccc}
    \toprule
    Integrated CDM & Metrics & M-M   & P-M   & PB-M \\
    \midrule
    \multirow{2}[2]{*}{KaNCD} & AUC (\%) & 81.34 & 81.39 & \textbf{81.49} \\
          & DOA (\%) & \textbf{79.59} & 78.92 & 78.62 \\
    \bottomrule
    \end{tabular}}
  \label{appd:fig:scale1}%
\end{table}

\begin{table}[!t]
  \centering
  \caption{Scaling up in training datasets to predict SLP-CHI.}
    \resizebox{0.75\linewidth}{!}{\begin{tabular}{c|c|ccc}
    \toprule
    Integrated CDM & Metrics & H-C   & HE-C  & HEG-C \\
    \midrule
    \multirow{2}[2]{*}{KaNCD} & AUC (\%) & 81.11 & 80.68 & \textbf{81.49} \\
          & DOA (\%) & 75.38 & \textbf{75.43} & 75.20  \\
    \bottomrule
    \end{tabular}}
  \label{appd:fig:scale2}%
\end{table}%


\subsection{Hyperparameter Analysis}\label{sec:hyper}

\begin{table}[!t]
  \centering
  \caption{Comparison of LRCD with different \textbf{TEM}. Details are the same as Table~\ref{tab:exp_subject}.}
    \resizebox{0.85\linewidth}{!}{
    \begin{tabular}{c|c|cccc}
    \toprule
    Datasets & Metrics & Bert  & LLama & ada & 3-large \\
    \midrule
    \multirow{2}[1]{*}{PB-M} & AUC (\%) & 80.30 & 79.50 & 80.09 & \textbf{80.32} \\
          & DOA (\%) & 76.75 & 76.16 & 76.92 & \textbf{77.21} \\
    \midrule
    \multirow{2}[2]{*}{EHG-M} & AUC (\%) & 79.72 & 79.09 & \textbf{80.17} & 79.98 \\
          & DOA (\%) & 76.71 & 75.93 & 76.56 & \textbf{76.75} \\
    \midrule
    \multirow{2}[2]{*}{PB-C} & AUC (\%) & \textbf{83.78} & 83.16 & 83.71 & 83.55 \\
          & DOA (\%) & \textbf{76.25} & 75.33 & 75.93 & 75.48 \\
    \midrule
    \multirow{2}[1]{*}{EHG-C} & AUC (\%) & 83.17 & 81.37 & \textbf{83.41} & 83.30 \\
          & DOA (\%) & \textbf{75.72} & 74.17 & 75.44 & 75.18 \\
    \midrule
    \multirow{2}[1]{*}{EDM-MATH} & AUC (\%) & \textbf{80.06} & 79.05 & 80.15 & 79.99 \\
          & DOA (\%) & \textbf{77.38} & 76.10 & 76.81 & 76.93 \\
    \midrule
    \multirow{2}[1]{*}{MATH-EDM} & AUC (\%) & 78.52 & 77.55 & 78.96 & \textbf{79.46} \\
          & DOA (\%) & 76.77 & 76.45 & 76.81 & \textbf{77.14} \\
    \midrule
    \multirow{2}[1]{*}{MOOC-MAT} & AUC (\%) & 75.39 & 77.28 & \textbf{78.97} & 77.23 \\
          & DOA (\%) & 73.93 & 74.15 & 75.56 & \textbf{75.76} \\
    \midrule
    \multirow{2}[2]{*}{MATH-MOOC} & AUC (\%) & 82.79 & 81.26 & \textbf{84.17} & 82.28 \\
          & DOA (\%) & \textbf{79.22} & 77.81 & 78.43 & 75.58 \\
    \bottomrule
\end{tabular}
}
  \label{tab:hyper_tem}%
\end{table}%
\textbf{The Effect of \textbf{TEM}.}
After obtaining the TCP, we utilize advanced TEM to generate representations in the language space. Here, we employ four renowned TEMs (i.e., Bert~\cite{DevLin2019Bert}, OpenAI-ada, OpenAI-3-large~\cite{Radford2019Language}) for experiments following~\cite{tao2022slmrec, liu2022elimrec, shen2024pmg, sun2022enhancing, sheng2024AlphaRec, liu2024PreferDiff}. As shown in Table~\ref{tab:hyper_tem}, within LRCD, using either OpenAI-ada or OpenAI-3-large results in strong performance. We recommend these two options. Notably, Bert also achieves commendable results, making it suitable for resource-constrained scenarios.

\textbf{The Effect of the Integrated CDM.} As LRCD is model-agnostic, we can integrate any existing CDMs. In this study, we employ four renowned CDMs (i.e., MIRT~\cite{Sympson1978Mirt}, KSCD~\cite{Ma2022Kscd}, KaNCD~\cite{Wang2023Kancd}, OR-KaNCD~\cite{QLL2024kdd}) for our experiments.  As shown in Figure~\ref{fig:hyper_cdm} in Appendix, OR-KaNCD generally outperforms both KaNCD and KSCD. This shows that OR-KaNCD is not only effective in standard settings but also across various ZSCD settings. Therefore, we recommend OR-KaNCD as the default CDM.

\subsection{Diagnosis Report Analysis}\label{sec:case}
\begin{figure}[!t]
\centering
\begin{minipage}{0.49\linewidth}\centering
    \includegraphics[width=\textwidth]{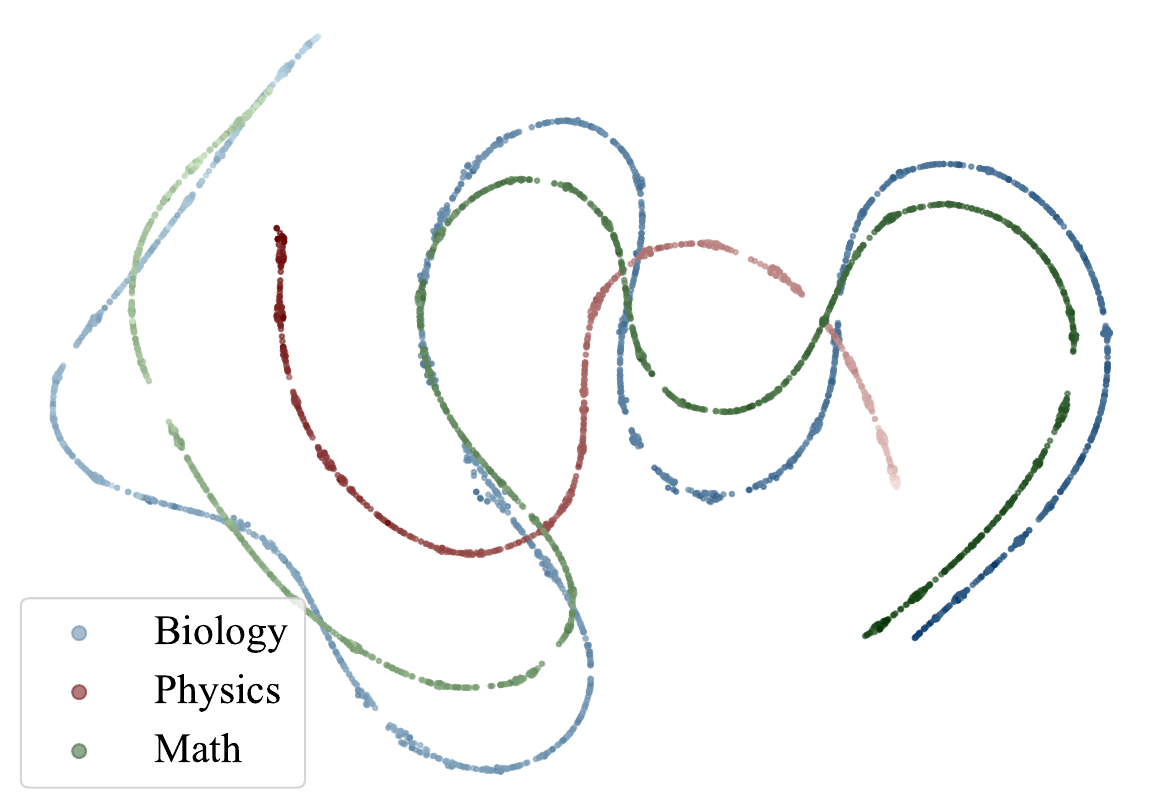}\\
    (a) Integrated CDM: KaNCD
\end{minipage}
\begin{minipage}{0.49\linewidth}\centering
     \includegraphics[width=\textwidth]{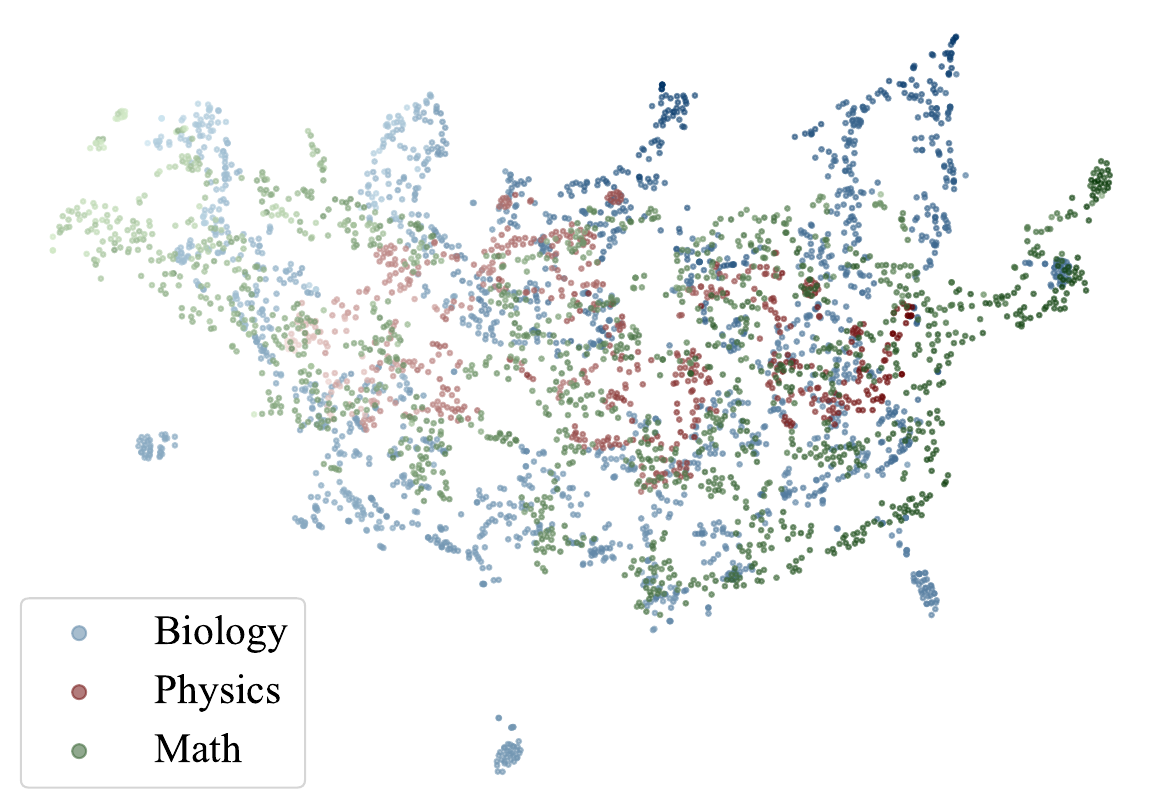}\\
     (b) Integrated CDM: OR-KaNCD
 \end{minipage}
\caption{t-SNE visualization of students' mastery levels in different domains.}
\label{fig:tsne}
\end{figure}

\textbf{Visualization of the Inferred Mastery Levels.} 
This subsection aims to further illustrate the implications of the students' mastery levels inferred by LRCD. We consider Biology and Physics as the source domains and Math as the target domain. Consequently, the inferred mastery levels of students in the source domains are obtained through supervised training, while those in the target domain are inferred without training. Following previous work~\cite{Wang2023Kancd}, we employ t-SNE~\cite{Van2009Tsne}, a renowned dimensionality reduction method, to map the mastery levels onto a two-dimensional plane. Firstly, we use different colors to represent different subjects. Then, we use shading to indicate the students' average correct rates, with darker shades representing higher correct rate. Finally, we visualize this using a scatter plot, as shown in Figure~\ref{fig:tsne}. We can observe that students with similar average correct rates tend to cluster together. Moreover, the mastery levels of students from different domains are well separated. This indicates that, although LRCD does not explicitly use methods to distinguish different domains and instead trains them together, it can still accurately differentiate students from various domains due to the distinct behavior patterns of each domain.

\begin{figure}[!t]
\centering
\includegraphics[width=0.98\linewidth]{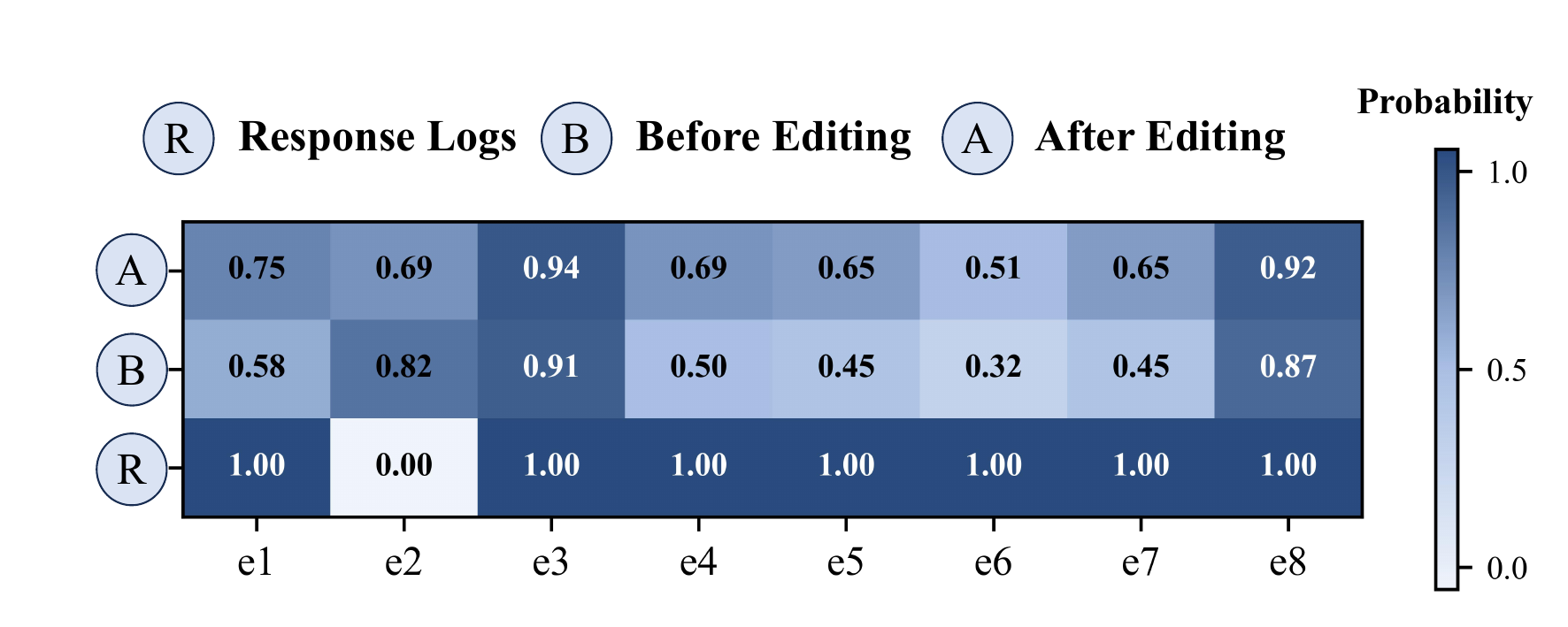}
\caption{Case study of student profile editing.}
\label{fig:case_study}
\end{figure}

\textbf{Student Profile Editing.} An additional advantage of LRCD is its capability to directly adjust students' mastery levels by editing their profiles, without necessitating data alteration and model retraining. Here, we select a student $s_i$ as an example who has not done any related exercises on the concept of ``Angle'' before editing. We give a detailed ID within the corresponding dataset in Appendix~\ref{appd:exp:case}. After LRCD training achieves convergence, we fix the parameters and obtain its current representation $\mathbf{h}_{s_i}^{(l)^-}$. We then acquire the student's new interactions in ``Angle'' and, using the method described in Section~\ref{sec:tcp}, derive the new representation $\mathbf{h}_{s_i}^{(l)^+}$. Finally, we combine the two representations in a 7:3 ratio and pass them through the student mapper to infer the student's new mastery level. To evaluate the accuracy of the diagnostic results, we assess whether the student's performance on Angle-related exercises has improved. As shown in Figure~\ref{fig:case_study}, the third row (i.e., Response Logs) indicates that this student answers all exercises correctly except for $e_2$. In the second row (i.e., Before Editing), we see the predictions given by LRCD before editing. It is evident that, although the model has not been trained on this student's interactions related to the ``Angle'', it still provides reasonable predictions. In the first row (i.e., After Editing), after editing, our predictions have all improved compared to the original ones. Notably, on exercise $e_2$, where the student initially got wrong, our prediction has become more accurate. This shows that LRCD can effectively adjust its assessment of a student's abilities through profile editing. This editing capability is highly beneficial in real educational scenarios, allowing students to preview their potential improvements in advance. By doing so, they can select appropriate exercises to focus on, thereby alleviating their academic burden.

\section{Conclusion}
This paper proposes a language representation favored zero-shot cross-domain cognitive diagnosis framework (LRCD) to address the limitations of existing CDMs, which often require specific models trained for specific domains. By leveraging textual descriptions to profile students, exercises, and concepts, LRCD transforms these profiles into vectors within a unified language space using advanced text-embedding modules. To bridge the gap between language space and cognitive diagnosis space, we introduce language-cognitive mappers in LRCD, enabling efficient integration and training with existing CDMs. Extensive experiments validate that LRCD achieves commendable zero-shot performance across different target domains and, in some instances, competes with classic CDMs trained on full response data. Notably, LRCD provides intriguing insight into the distinctions between various subjects and educational sources. However, while LRCD shows significant efficacy, further development of more interpretable methods is needed to fully elucidate the mapping process and enhance the framework's applicability in online intelligent education systems.


\begin{acks}
We would like to thank the anonymous reviewers for their constructive comments. We also would like to thank Xinyue Ma for the reliable help. The algorithms and datasets in the paper do not involve any ethical issue. This work is supported by the National Natural Science Foundation of China (No. 62476091, No. 62106076).

\end{acks}

\newpage
\bibliographystyle{ACM-Reference-Format}
\balance
\bibliography{reference}
\newpage
\appendix
\section*{Appendix}


The appendix is organized as follows:


$\bullet$ Appendix~\ref{appd:time} presents a detailed comparison of training time and inference time with other baselines.

$\bullet$ Appendix~\ref{appd:exp} provides additional details on the experiments carried out in this paper, including the details about datasets, the explanation of DOA, implementation of baselines, experimental results in the standard setting, as well as details on hyperparameter analysis and case studies.

\section{Time Comparison} \label{appd:time}
Here, we use Physics and Biology as the source domains, and Math as the target domain, as an example to explore the comparison of different models in terms of training time and inference time.

\subsection{Training Time}\label{appd:time:train}
\begin{figure}[!htbp]
    \centering
    \begin{minipage}[b]{0.48\linewidth}
        \centering
        \includegraphics[width=\linewidth]{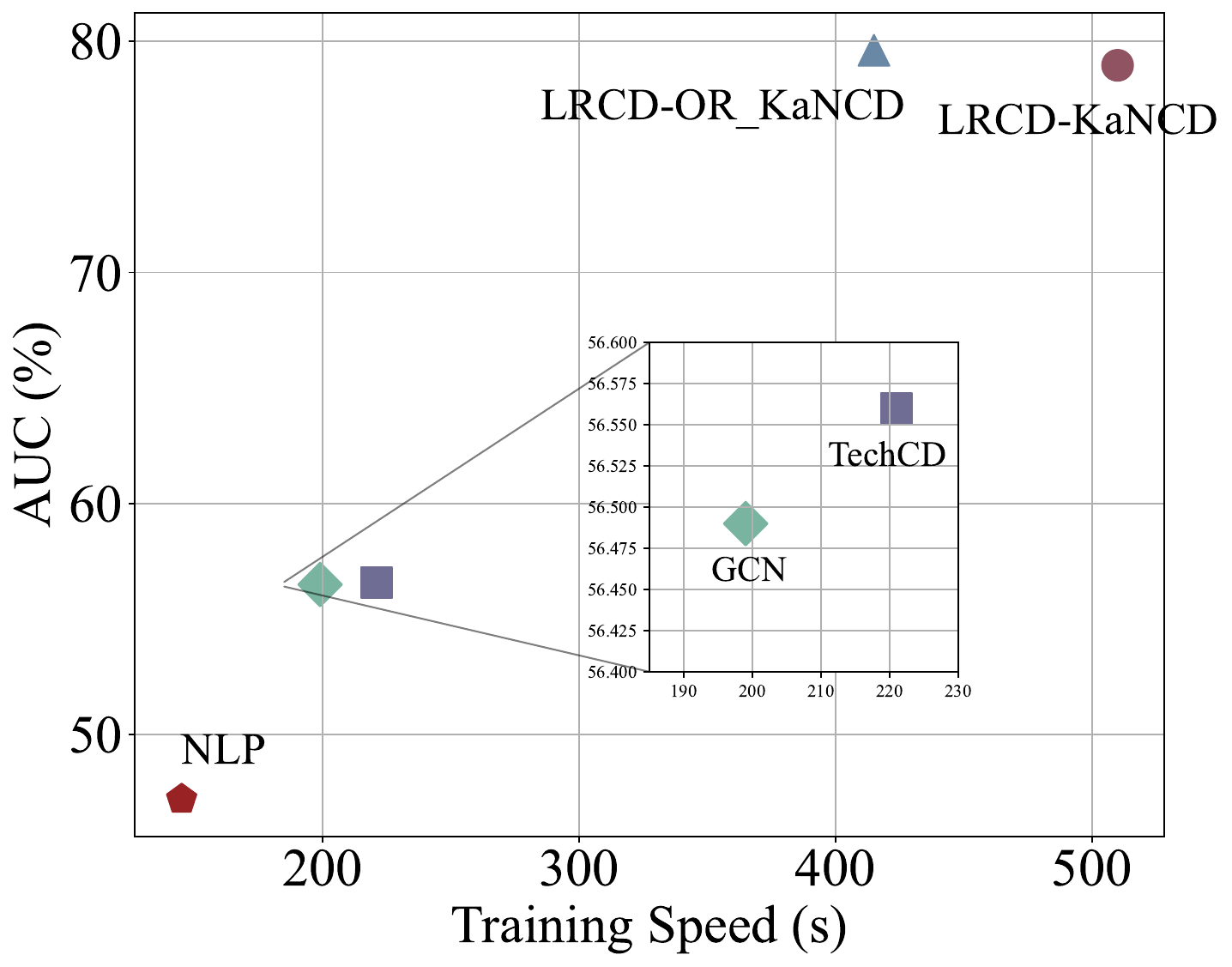}
        \\
        (a)
        \label{fig:infer}
    \end{minipage}
    \hfill
    \begin{minipage}[b]{0.48\linewidth}
        \centering
        \includegraphics[width=\linewidth]{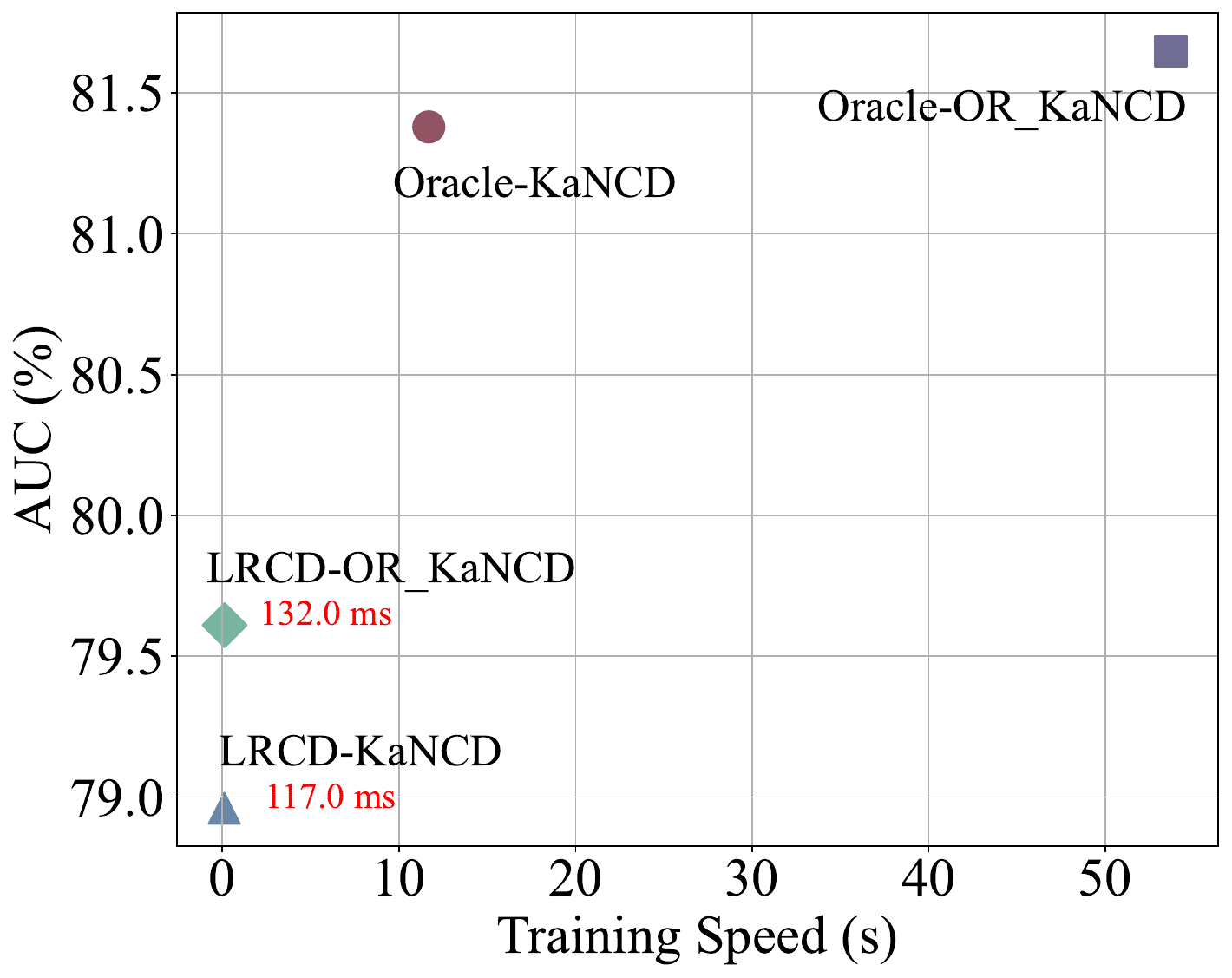} \\
        (b)
        \label{fig:train}
    \end{minipage}
    \caption{(a) Training time comparison with baselines. (b) Inference time comparison with the model retraining}
    \label{fig:combined}
\end{figure}

As discussed in Section~\ref{sec:discuss}, by employing a space-for-time strategy, we can store the obtained text embeddings locally, thereby minimizing additional time requirements. Here, we provide the actual running time of LRCD compared with the baselines in ZSCD. Note that the training time here is the time it takes for the models to reach the optimal AUC. As shown in Figure~\ref{fig:combined}(a), we select KaNCD and OR-KaNCD for our model. It can be seen that although our model takes longer time, it has fine predictive performance compared to other baselines.

\subsection{Inference Time}\label{appd:time:infer}

As we claim in the introduction, when new domain's response logs emerges, OIDP often need to leverage prior knowledge to quickly and accurately provide diagnostic results for students without retraining models. Therefore, inference time, which equates to the system's response time to the user, is crucial. Here, we compare the inference time with the model retraining time.

As shown in Figure~\ref{fig:combined}(b), we selected KaNCD and OR-KaNCD as diagnostic methods. For KaNCD, the diagnostic time of our model is 99 times longer than that of the oracle, while OR-KaNCD is 406 times longer. This indicates that our model has good time utility while maintaining great diagnostic performance, and can quickly diagnose ``zero sample'' students.

\section{Experiments}\label{appd:exp}
\subsection{Details about the Datasets}\label{appd:exp:dataset}
Here, we provide detailed information regarding the SLP dataset, including the number of students and exercises for each subject in Table~\ref{tab:dataset_slp}. The specific details align with those presented in Table~\ref{tab:dataset_1}.

$\bullet$ \textbf{SLP}~\cite{Lu2021slp}: The SLP dataset for K-12 education encompasses five dimensions: student demographics, psychometric intelligence, academic performance, family, and school information. It automatically records students' academic performance in eight subjects (Math, Physics, Chemistry, Biology, History, Chinese, Geography, English) over three years (7th to 9th grade).

$\bullet$ \textbf{MOOC}~\cite{Yu2023MOOCRadar}: The MOOC dataset supports cognitive student modeling in Massive Open Online Courses by providing learning resources, structures, and content related to students’ exercise behaviors. It also includes Chinese contextual information for exercises and concepts.

$\bullet$ \textbf{EDM}~\cite{edm-cup-2023}: Originating from the EDM Cup 2023 competition, the EDM dataset focuses on predicting students’ end-of-unit assignment scores using their click-stream data from previous in-unit assignments on the ASSISTments platform. It contains millions of student actions along with detailed information on the curricula, assignments, problems, and the tutoring provided.

\begin{table}[!htbp]
  \centering
  \caption{Details about different subjects in SLP.}
    \resizebox{\linewidth}{!}{\begin{tabular}{l|ccccccc}
    \toprule
    Datasets & Math  & Chinese & Physics & Biology & English & History & Geography \\
    \midrule
    \#Students & 1,475  & 623   & 639   & 1,940  & 306   & 1,603  & 1,077 \\
    \#Exercises & 615   & 510   & 1,441  & 773   & 355   & 752   & 427 \\
    \#Concepts & 33    & 17    & 50    & 16    & 18    & 20    & 25 \\
    \#Response logs & 55,332 & 29,202 & 37,405 & 81,838 & 3,872  & 63,207 & 28,535 \\
    Average Correct Rate  & 0.550 & 0.588 & 0.611 & 0.503 & 0.366 & 0.418 & 0.371 \\
    Q Density  & 1,000 & 1.000 & 1.000 & 1.000 & 1.000 & 1.000 & 1.002 \\
    Category& Science & Humanity & Science & Science & Humanity & Humanity & Humanity \\
    \bottomrule
    \end{tabular}}
  \label{tab:dataset_slp}
\end{table}

\subsection{Degree of Agreement (DOA)}\label{appd:exp:doa}
Here, we provide a further explanation regarding the degree of agreement. Suppose that the inferred students' mastery levels are represented by $\textbf{Mas} \in \mathbb{R}^{N \times K}$, where $N$ denotes the number of students and $K$ signifies the number of concepts. The underlying intuition here is that if the student $s_a$ demonstrates higher accuracy in answering exercises related to the concept $c_k$ compared to the student $s_b$, then the probability of $s_a$ mastering $c_k$ should be greater than that of $s_b$. In other words, $\mathbf{Mas}_{s_a, c_k} > \mathbf{Mas}_{s_b, c_k}$. The Degree of Agreement (DOA) is defined as in Eq.~\eqref{eq:DOA}
\begin{equation}\label{eq:DOA}
    \resizebox{0.9\linewidth}{!}{
    $\text{DOA}_{k}=\frac{1}{Z}\sum\limits_{a, b \in S} \delta\left(\textbf{Mas}_{s_a, c_k}, \textbf{Mas}_{s_b, c_k}\right) \frac{\sum_{j=1}^{M} \mathbf{Q}_{j k} \wedge \varphi(j, a, b) \wedge \delta\left(r_{a j}, r_{b j}\right)}{\sum_{j=1}^{M} \mathbf{Q}_{j k} \wedge \varphi(j, a, b) \wedge I\left(r_{a j} \neq r_{b j}\right)} \,,$
    }
\end{equation}
where $Z=\sum_{a, b \in S}\delta(\textbf{Mas}_{s_a, c_k}, \textbf{Mas}_{s_b, c_k})$, $\mathbf{Q}_{jk}$ indicates exercise $e_j$'s relevance to concept $c_k$, $\varphi(j, a, b)$ verifies if both students $s_a$ and $s_b$ answered $e_j$, $r_{aj}$ represents the response of $s_a$ to $e_j$, and $I (r_{a j} \neq r_{b j})$ checks if their responses are different, $\delta(r_{a j}, r_{b j})$ is $1$ for a correct response by $s_a$ and an incorrect response by $s_b$, and $0$ otherwise.

\begin{figure*}[!t]
\centering
\begin{minipage}{0.20\linewidth}\centering
    \includegraphics[width=\textwidth]{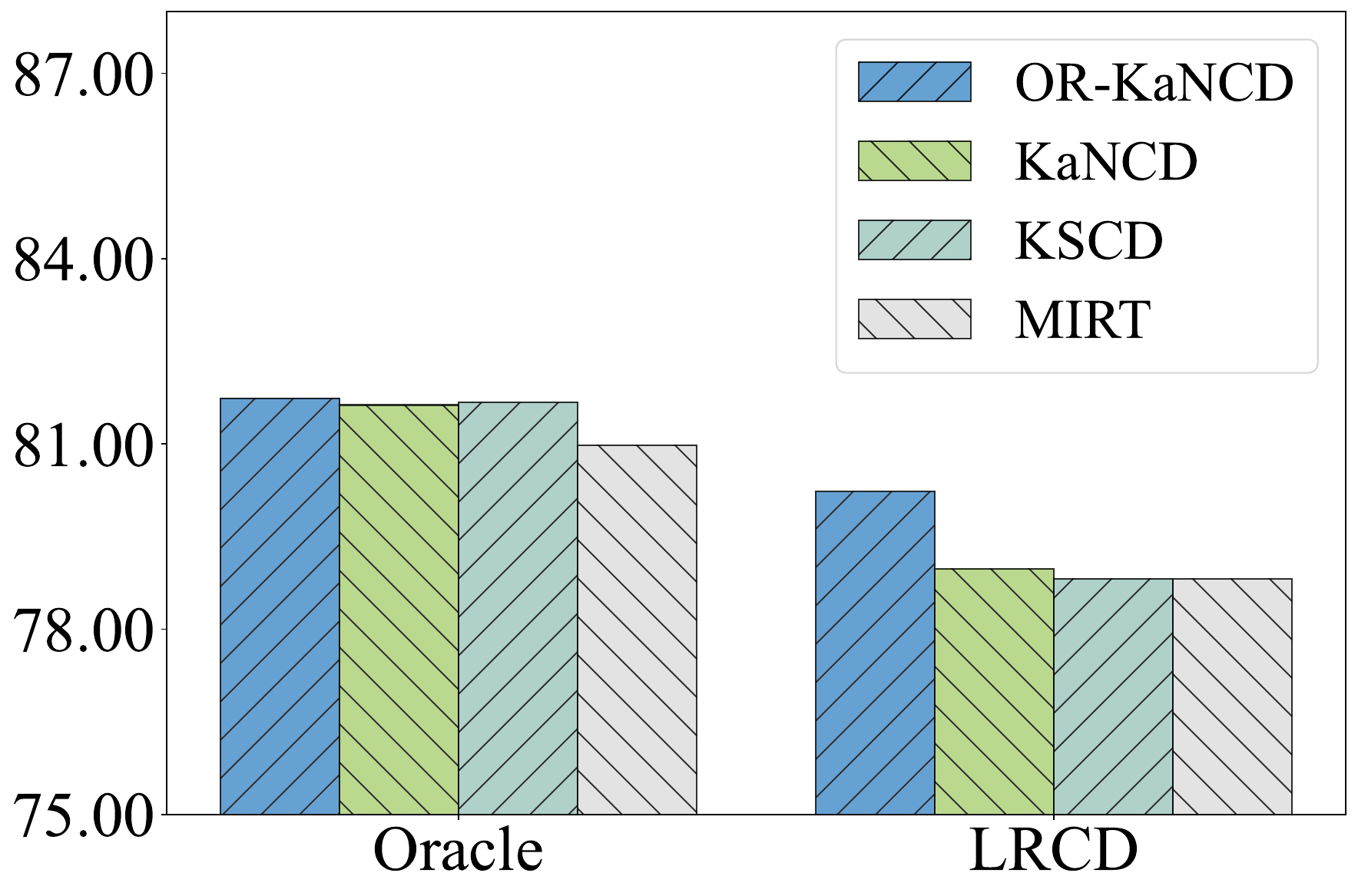}\\
    (a) PB-M
\end{minipage}
\begin{minipage}{0.20\linewidth}\centering
    \includegraphics[width=\textwidth]{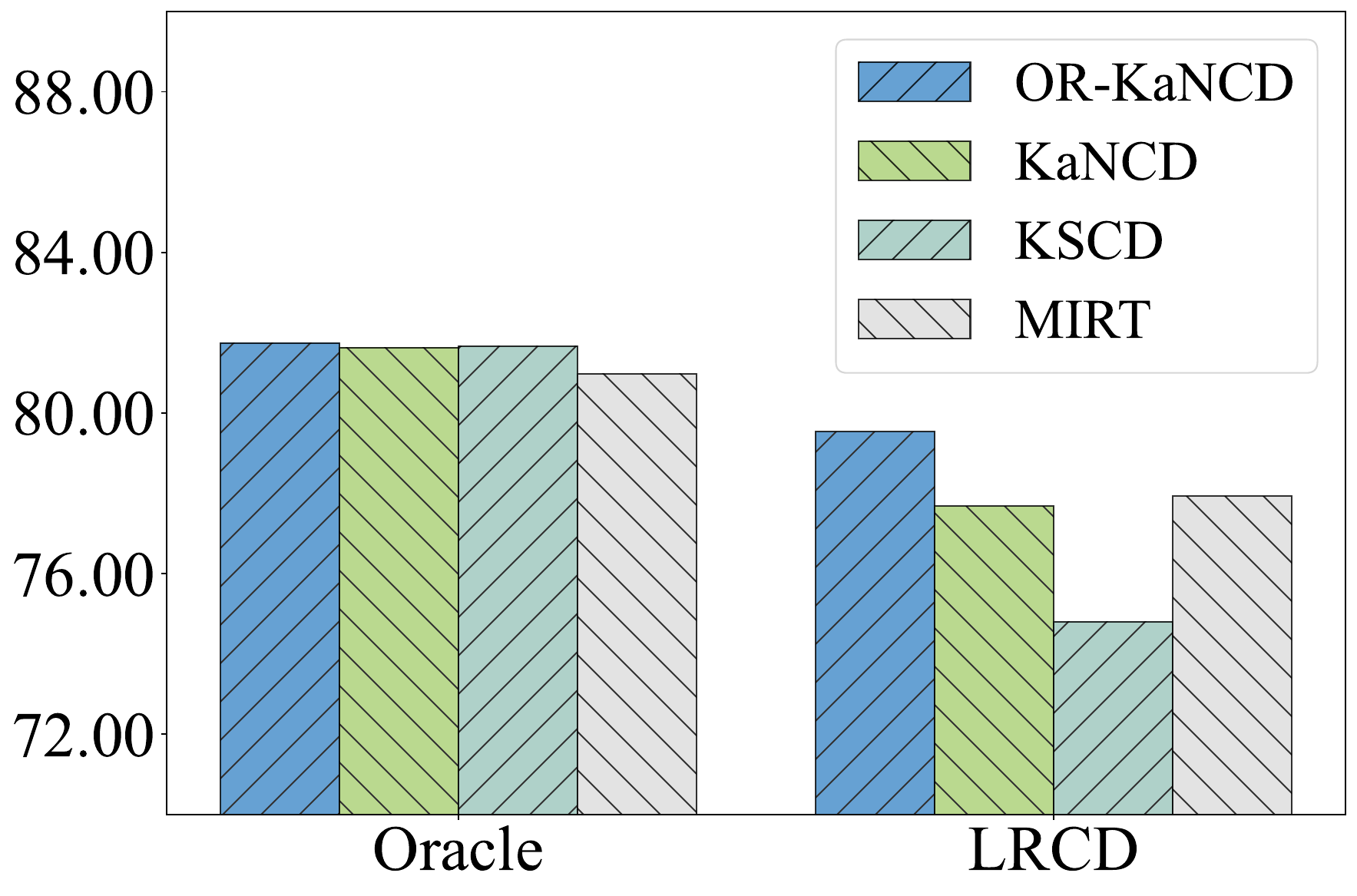}\\
    (b) EHG-M
\end{minipage}
\begin{minipage}{0.20\linewidth}\centering
    \includegraphics[width=\textwidth]{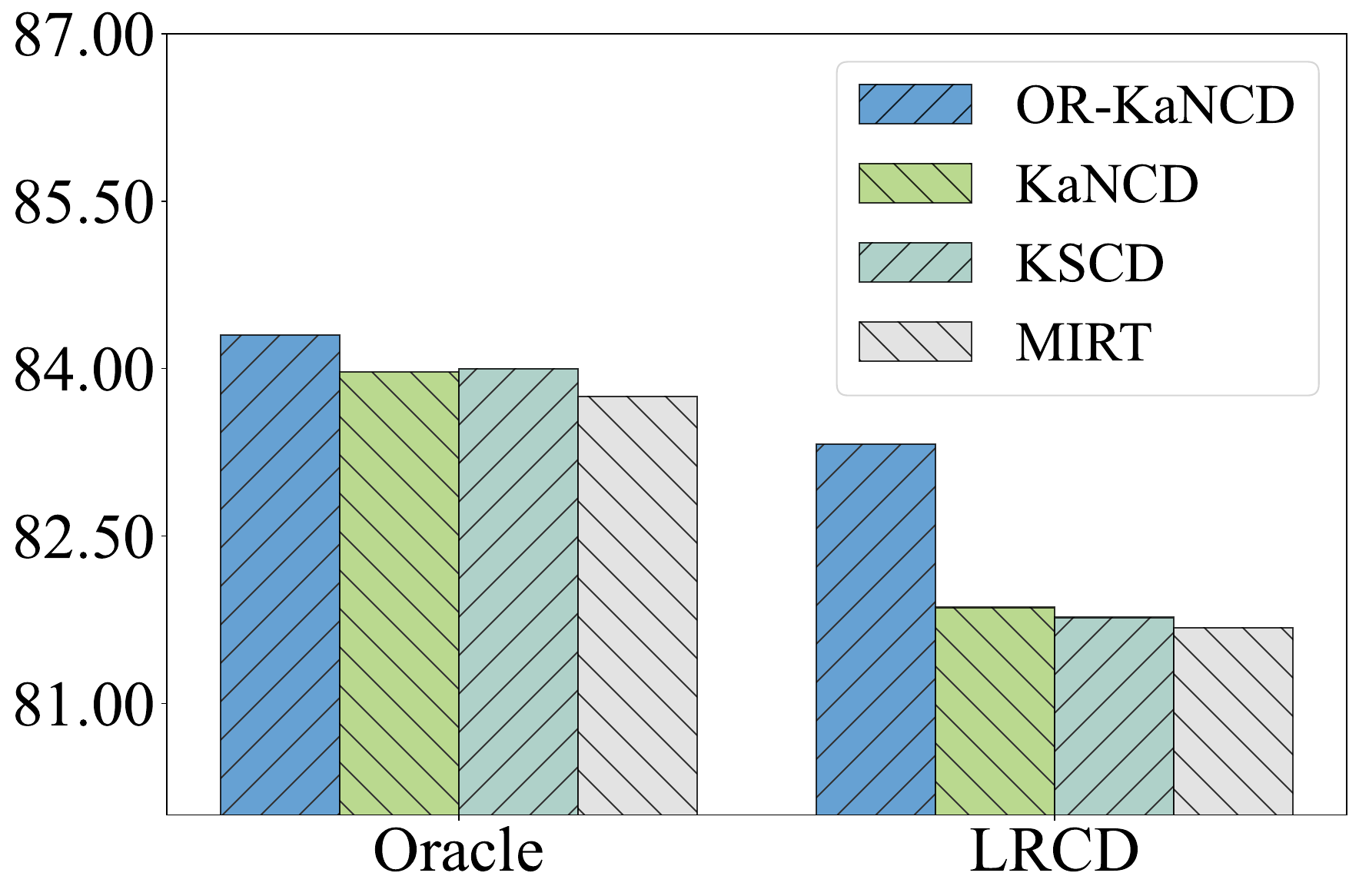}\\
    (c) PM-C
\end{minipage}
\begin{minipage}{0.20\linewidth}\centering
    \includegraphics[width=\textwidth]{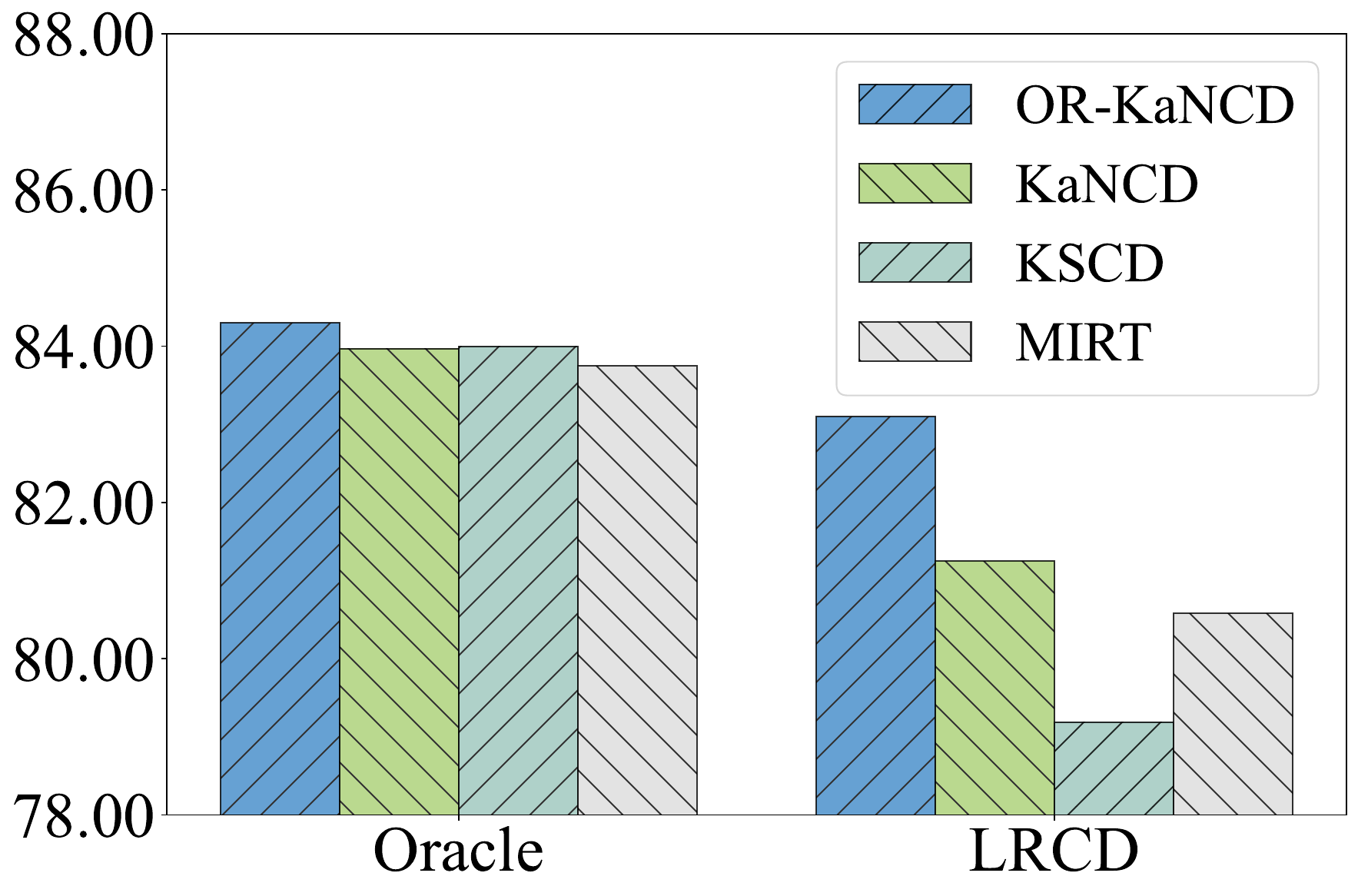}\\
    (d) EHG-C
\end{minipage}\\
\begin{minipage}{0.20\linewidth}\centering
    \includegraphics[width=\textwidth]{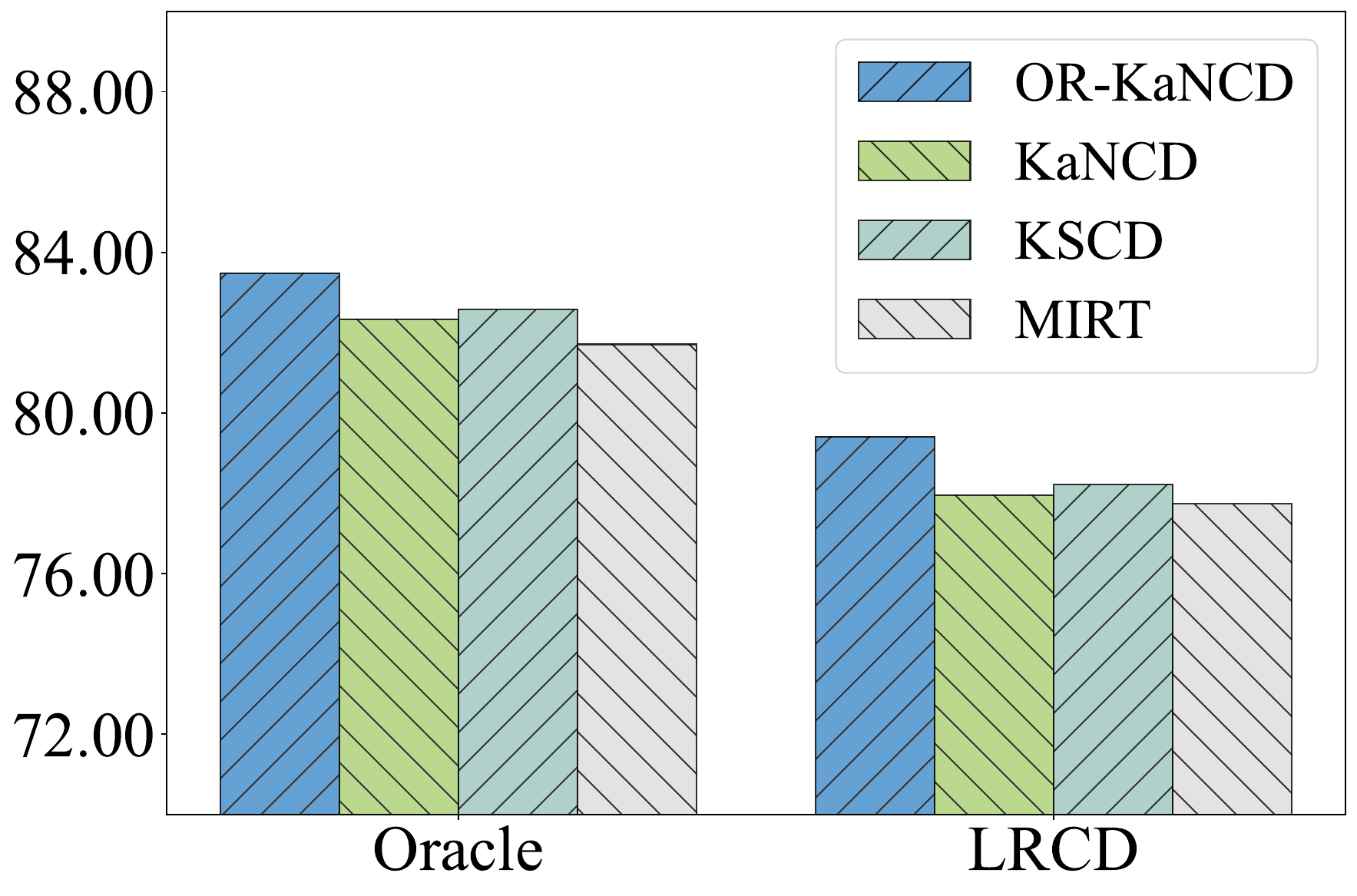}\\
    (a) Math-EDM
\end{minipage}
\begin{minipage}{0.20\linewidth}\centering
    \includegraphics[width=\textwidth]{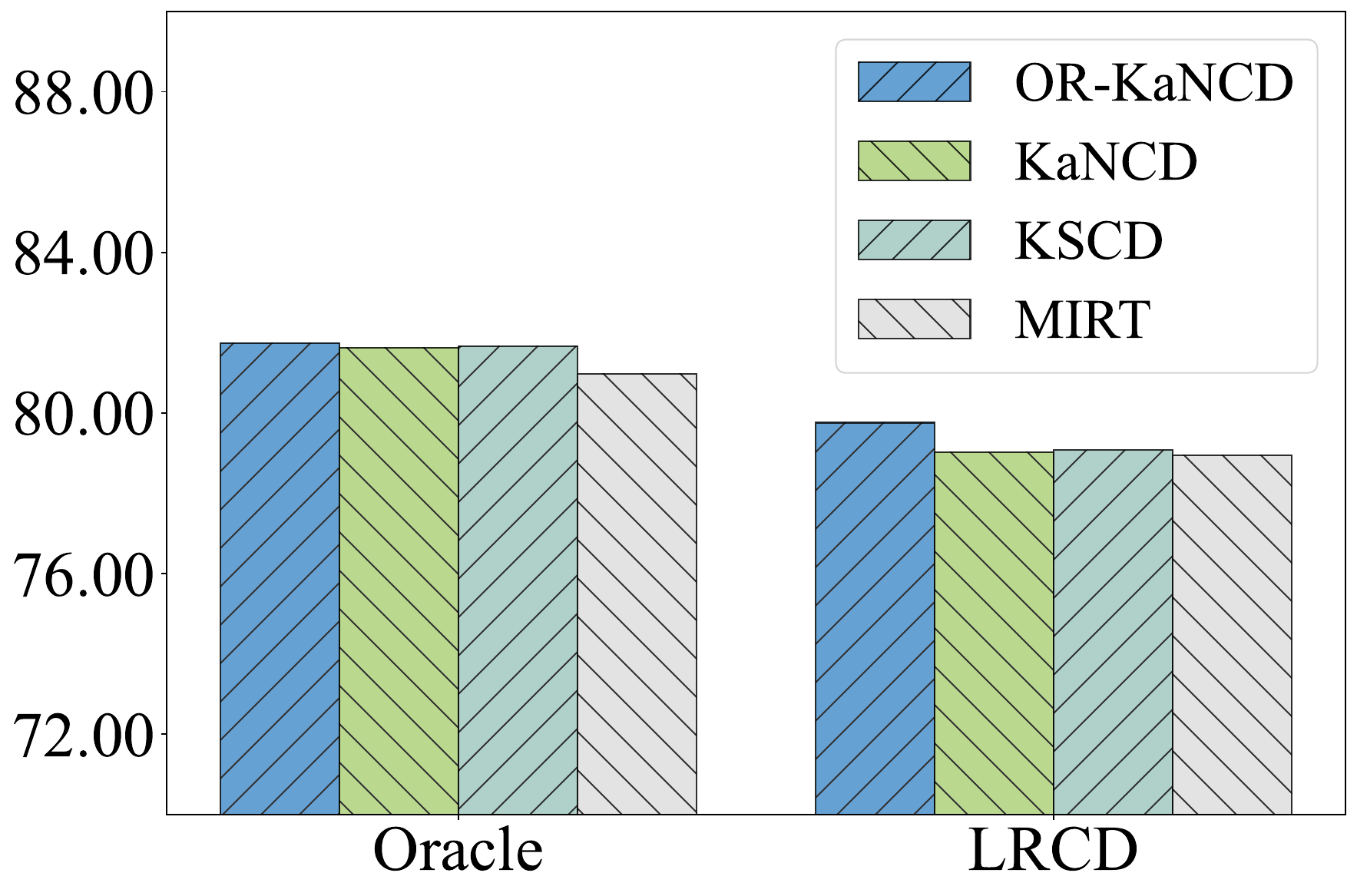}\\
    (b) EDM-Math
\end{minipage}
\begin{minipage}{0.20\linewidth}\centering
    \includegraphics[width=\textwidth]{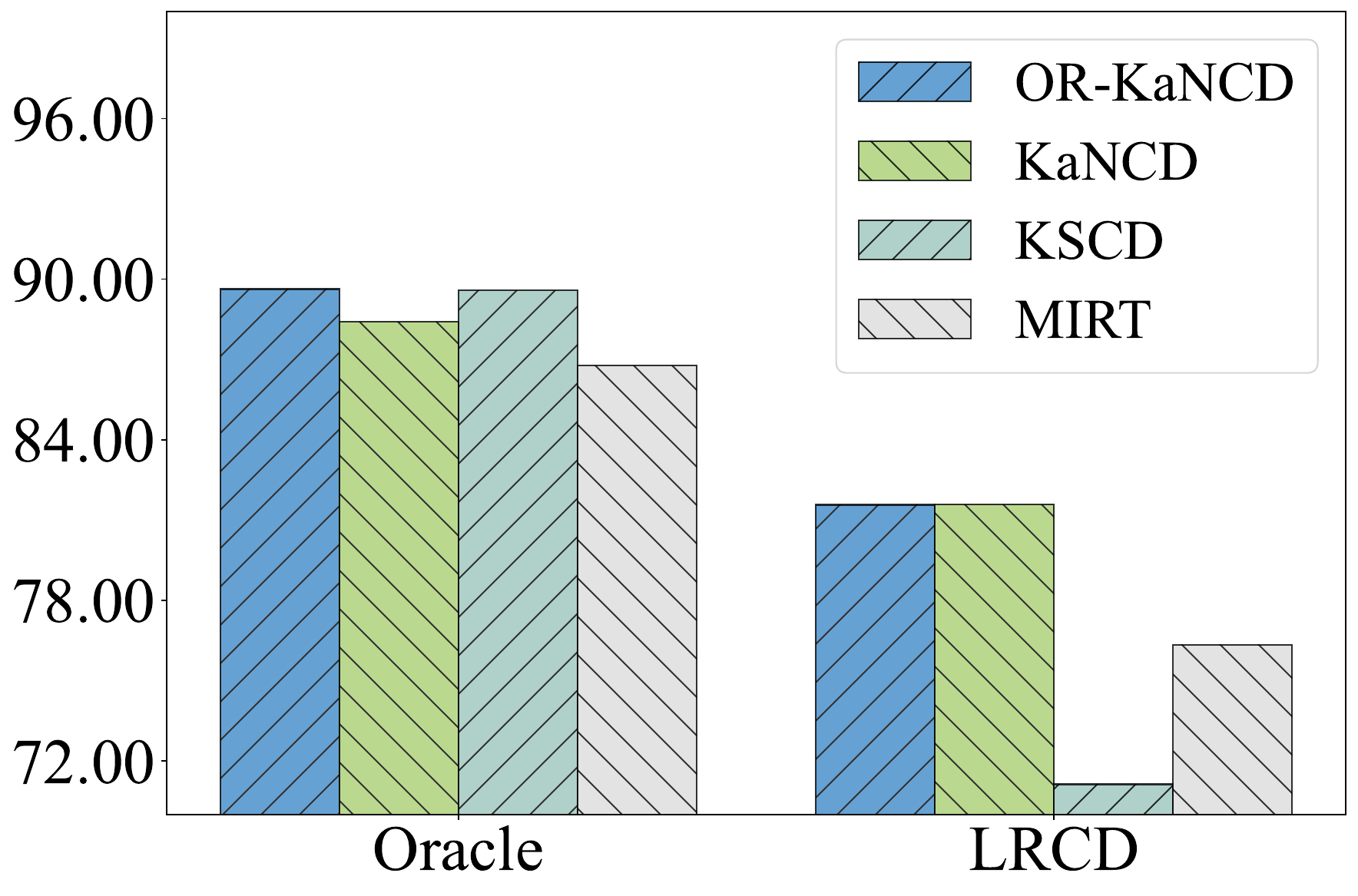}\\
    (c) Math-MOOC
\end{minipage}
\begin{minipage}{0.20\linewidth}\centering
    \includegraphics[width=\textwidth]{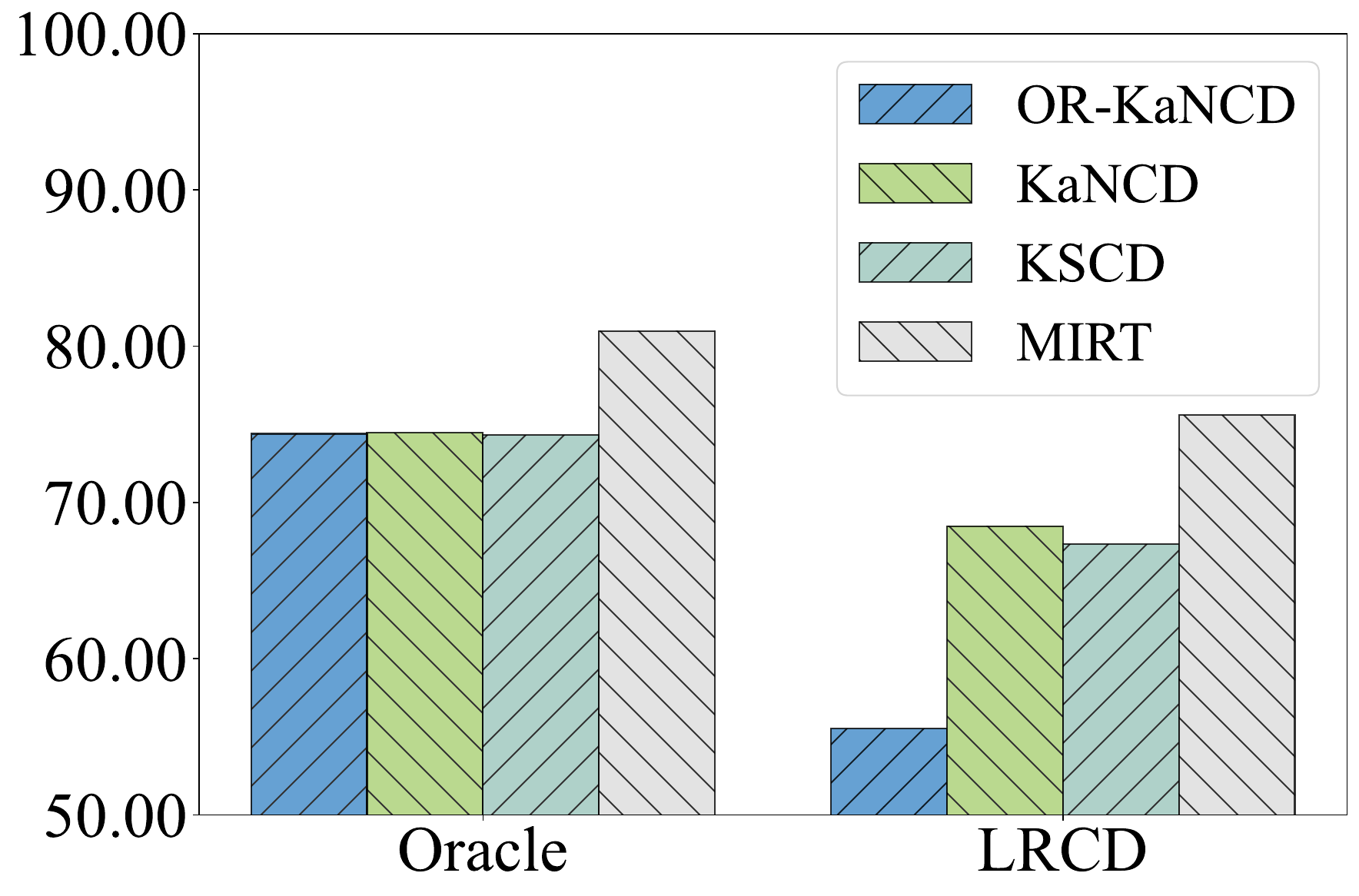}\\
    (d) MOOC-Math
\end{minipage}
\caption{Comparison of LRCD with different integrated CDM.}
\label{fig:hyper_cdm}
\end{figure*}

\subsection{Details about Baselines}\label{appd:exp:baselines}
In the following, we elaborate on some details regarding the utilization of the compared methods. 

 $\bullet$ KSCD~\cite{Ma2022Kscd} explores the implicit association among concepts and leverages a knowledge-enhanced interaction function. 

$\bullet$ KaNCD~\cite{Wang2023Kancd} enhances NCDM by investigating the implicit associations among concepts to address the issue of knowledge coverage. Here, we adopt the default parameters in the paper. 

$\bullet$ TechCD~\cite{Gao2023TechCD} uses a pedagogical knowledge concept graph as a mediator to connect students in the source domain with those in the target domain, thus effectively transferring student cognitive signals from source domains to target domains. Following~\cite{Gao2024Z13}, we utilize the statistical method proposed in~\cite{Gao2021Rcd} to construct the graph.

$\bullet$ Zero-1-3~\cite{Gao2024Z13} uses dual regularizers to split student embeddings into domain-shared and domain-specific components. It generates simulated practice logs for new target domain students by analyzing early-bird behaviors. Since exercise content is unavailable, we create exercise texts as described in Section~\ref{sec:tcp} and embed them using BERT~\cite{DevLin2019Bert}. For early bird students, we randomly select 10\% of the target domain due to the limited number of students.

 The implementation of MIRT, KaNCD comes from the public repository \url{https://github.com/bigdata-ustc/EduCDM}. For KSCD, we adopt the implementation from the authors in \url{https://github.com/BIMK/Intelligent-Education/tree/main/KSCD_Code_F}. For OR-KaNCD, we adopt the implementation from the authors in \url{https://github.com/ECNU-ILOG/ORCDF}. For TechCD, we adopt the implementation from the authors in \url{https://github.com/bigdata-ustc/TechCD}. 
For Zero-1-3, since there was no publicly available code when we submitted this paper, we have implemented it ourselves.

\subsection{Standard Setting}\label{appd:exp:standard}
\begin{figure}[!t]
\centering
\begin{minipage}{0.32\linewidth}\centering
    \includegraphics[width=0.8\textwidth]{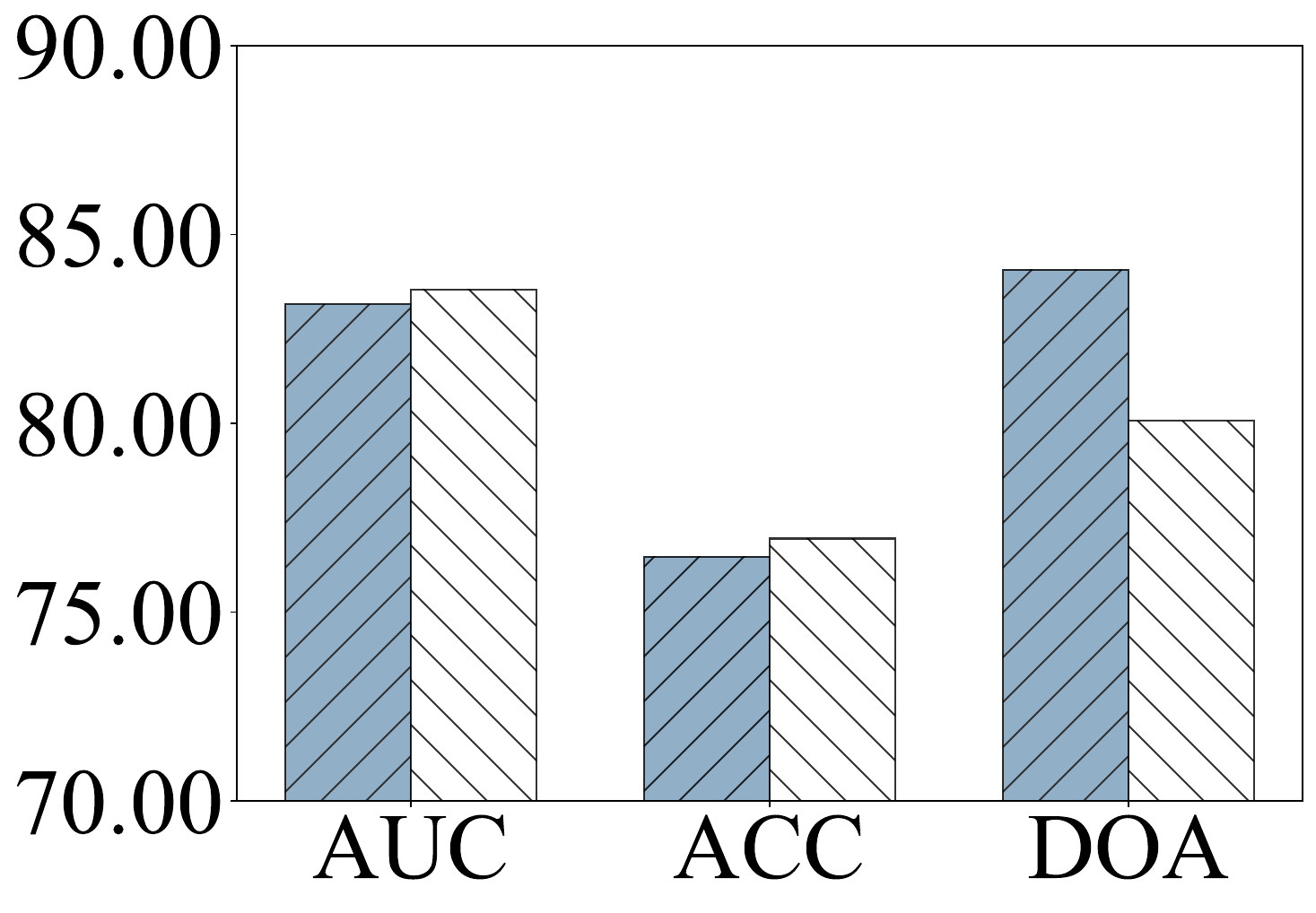}\\
    (a) EDM
\end{minipage}
\begin{minipage}{0.32\linewidth}\centering
    \includegraphics[width=0.8\textwidth]{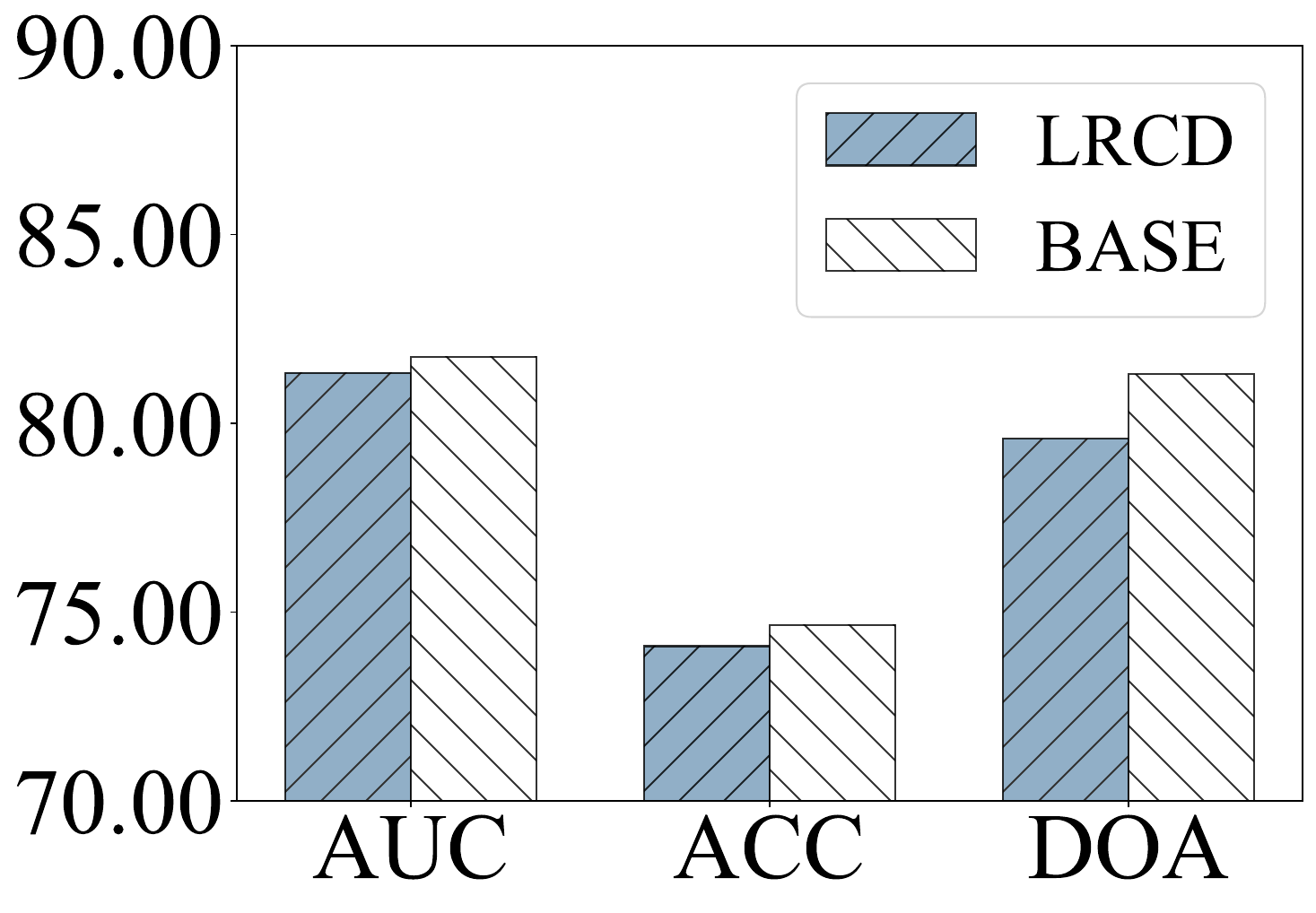}\\
    (b) SLP-Math
\end{minipage}
\begin{minipage}{0.32\linewidth}\centering
    \includegraphics[width=0.8\textwidth]{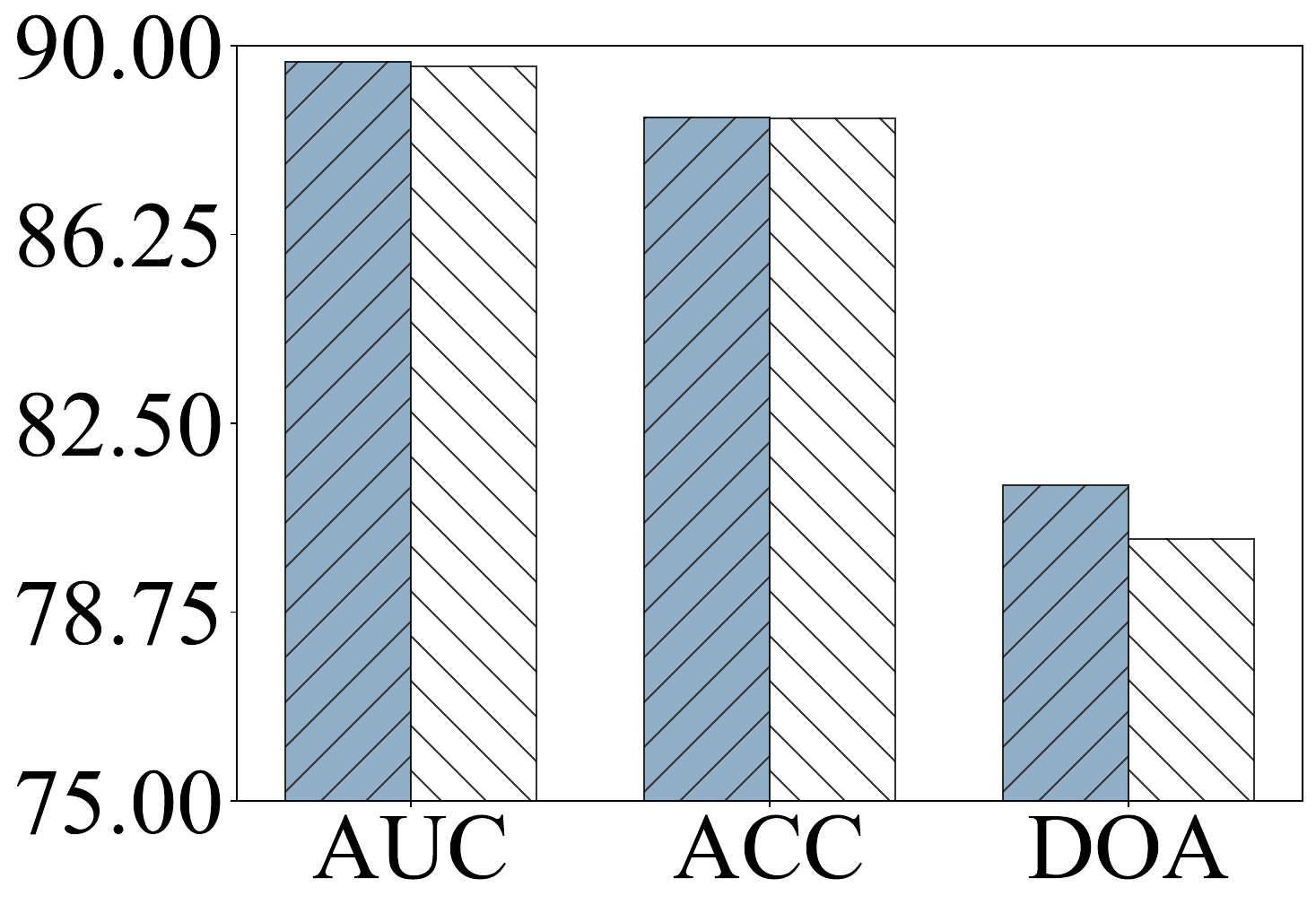}\\
    (c)  MOOC
\end{minipage}
\caption{Comparison with OR-KaNCD in the standard setting.}
\label{fig:exp:standard_or_kancd}
\end{figure}

\begin{figure}[!t]
\centering
\begin{minipage}{0.32\linewidth}\centering
    \includegraphics[width=0.8\textwidth]{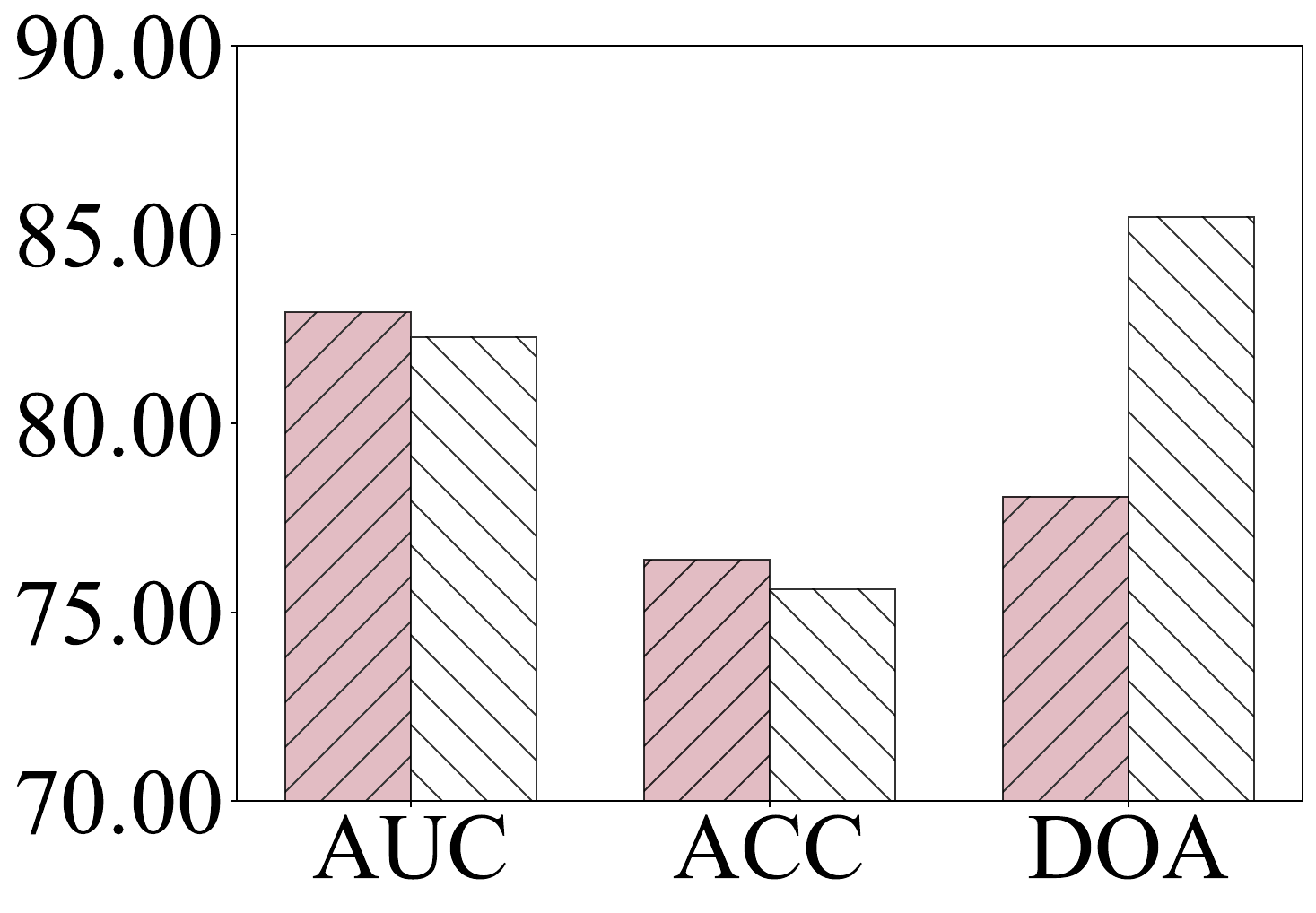}\\
    (a) EDM
\end{minipage}
\begin{minipage}{0.32\linewidth}\centering
    \includegraphics[width=0.8\textwidth]{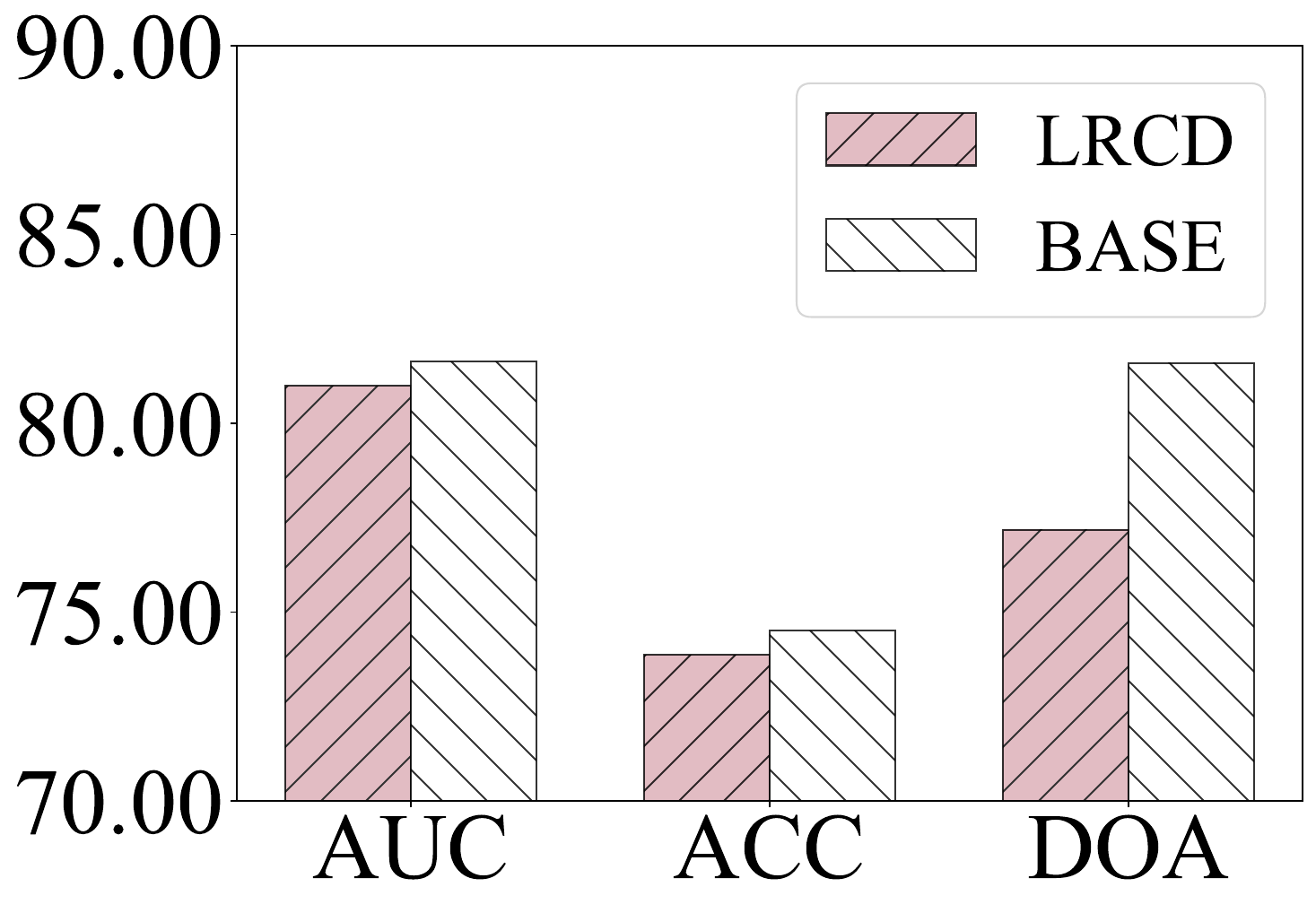}\\
    (b) SLP-Math
\end{minipage}
\begin{minipage}{0.32\linewidth}\centering
    \includegraphics[width=0.8\textwidth]{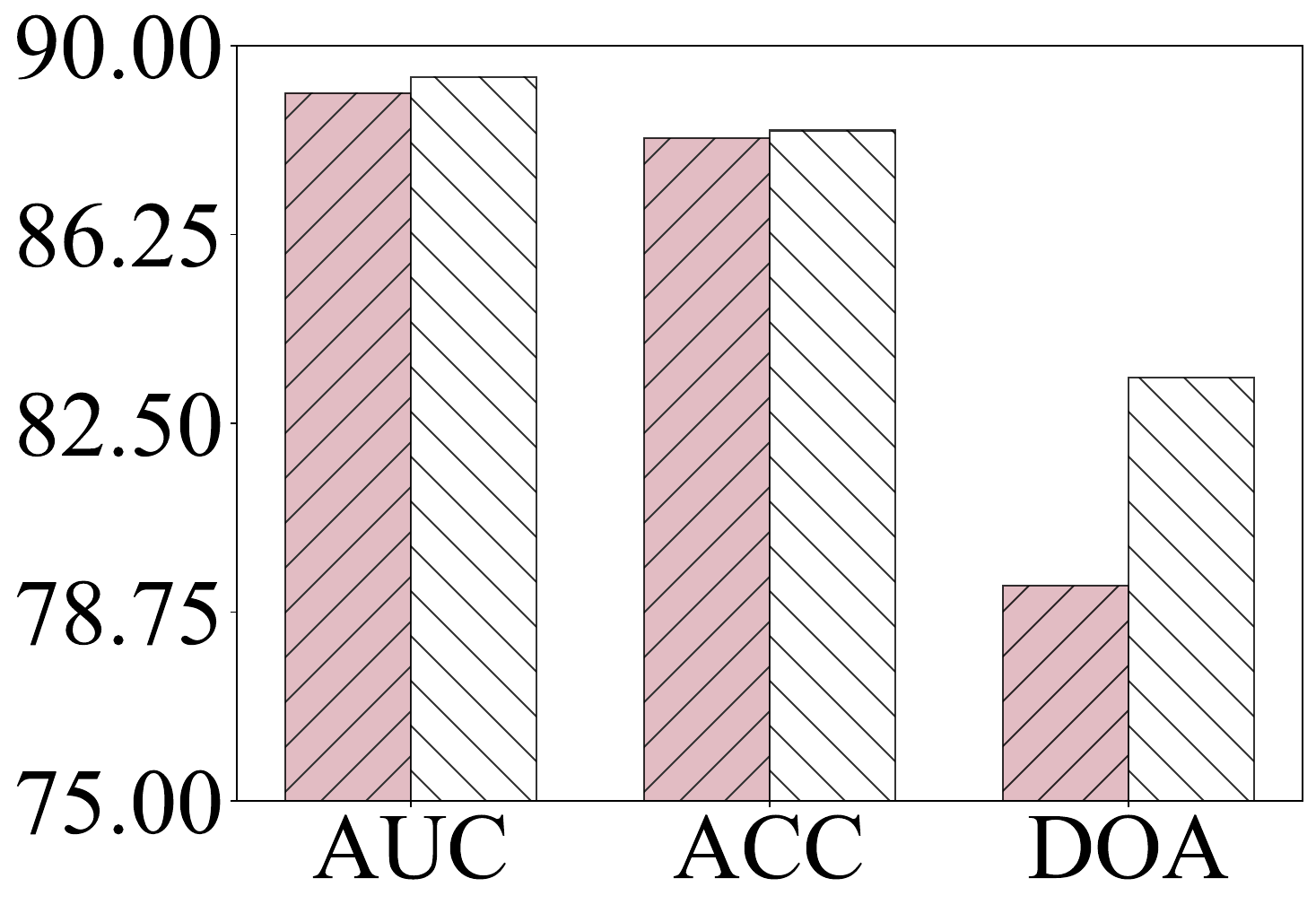}\\
    (c)  MOOC
\end{minipage}
\caption{Comparison with KaNCD in the standard setting.}
\label{fig:exp:standard_kancd}
\end{figure}

\begin{figure}[!t]
\centering
\begin{minipage}{0.32\linewidth}\centering
    \includegraphics[width=0.8\textwidth]{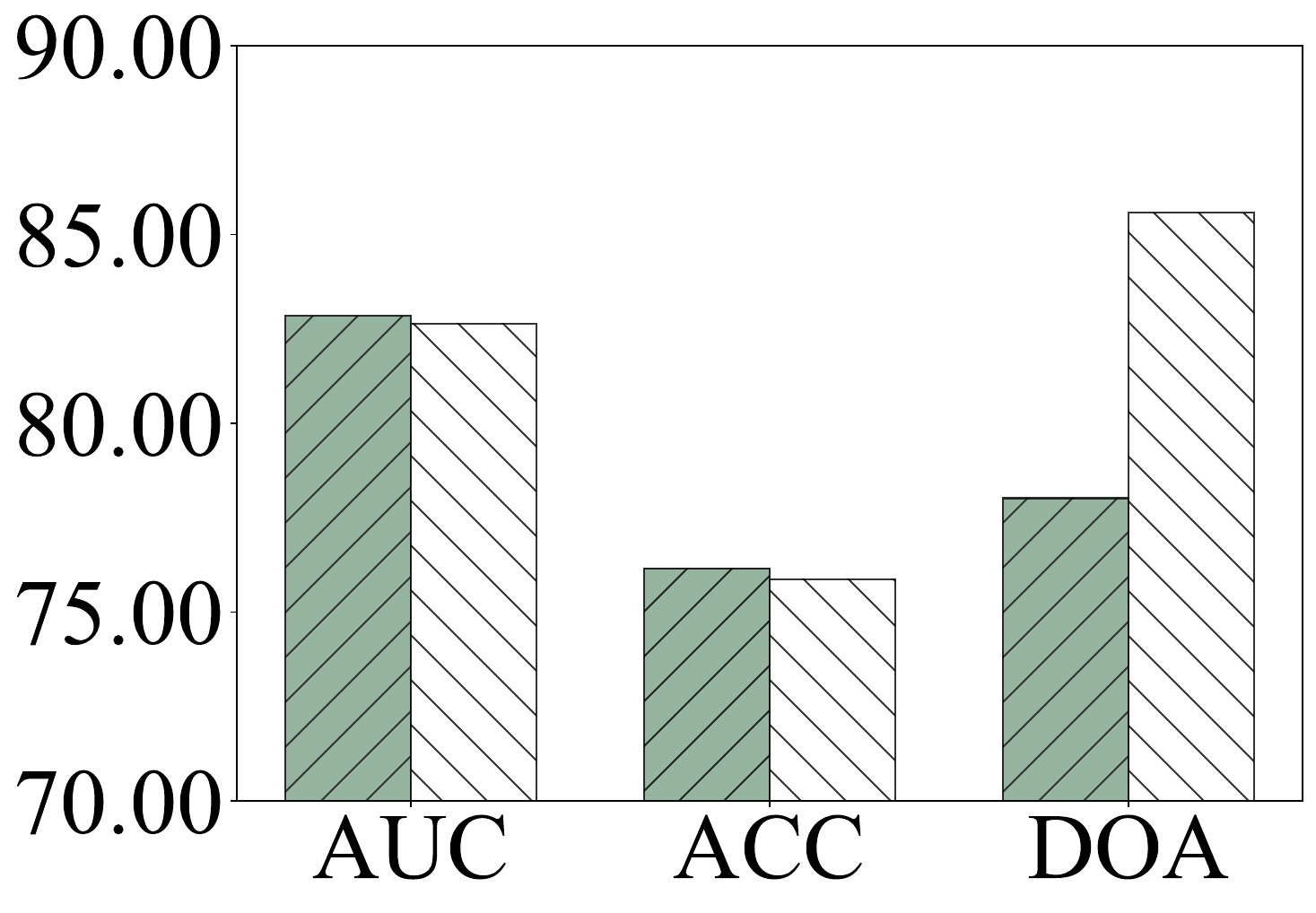}\\
    (a) EDM
\end{minipage}
\begin{minipage}{0.32\linewidth}\centering
    \includegraphics[width=0.8\textwidth]{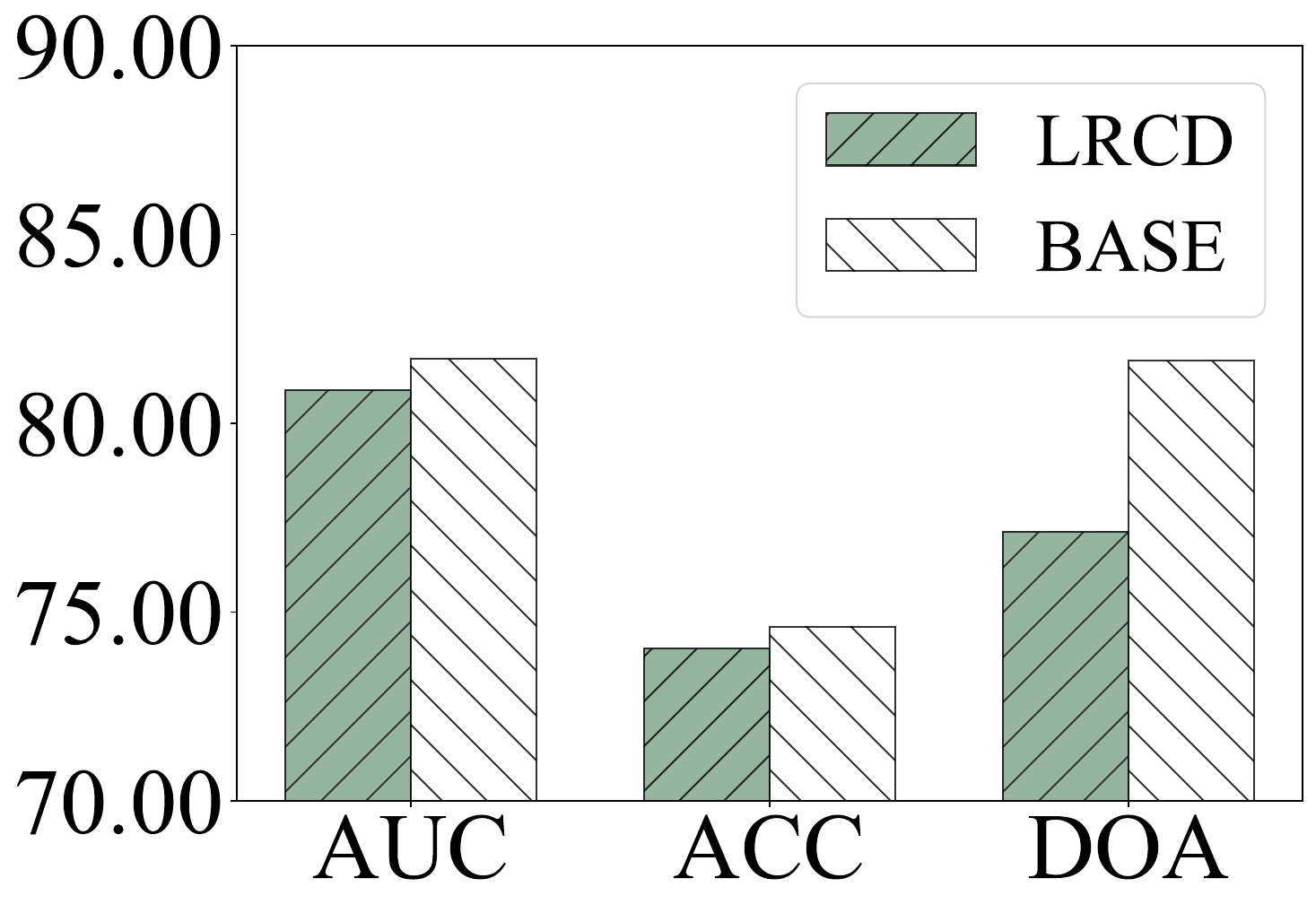}\\
    (b) SLP-Math
\end{minipage}
\begin{minipage}{0.32\linewidth}\centering
    \includegraphics[width=0.8\textwidth]{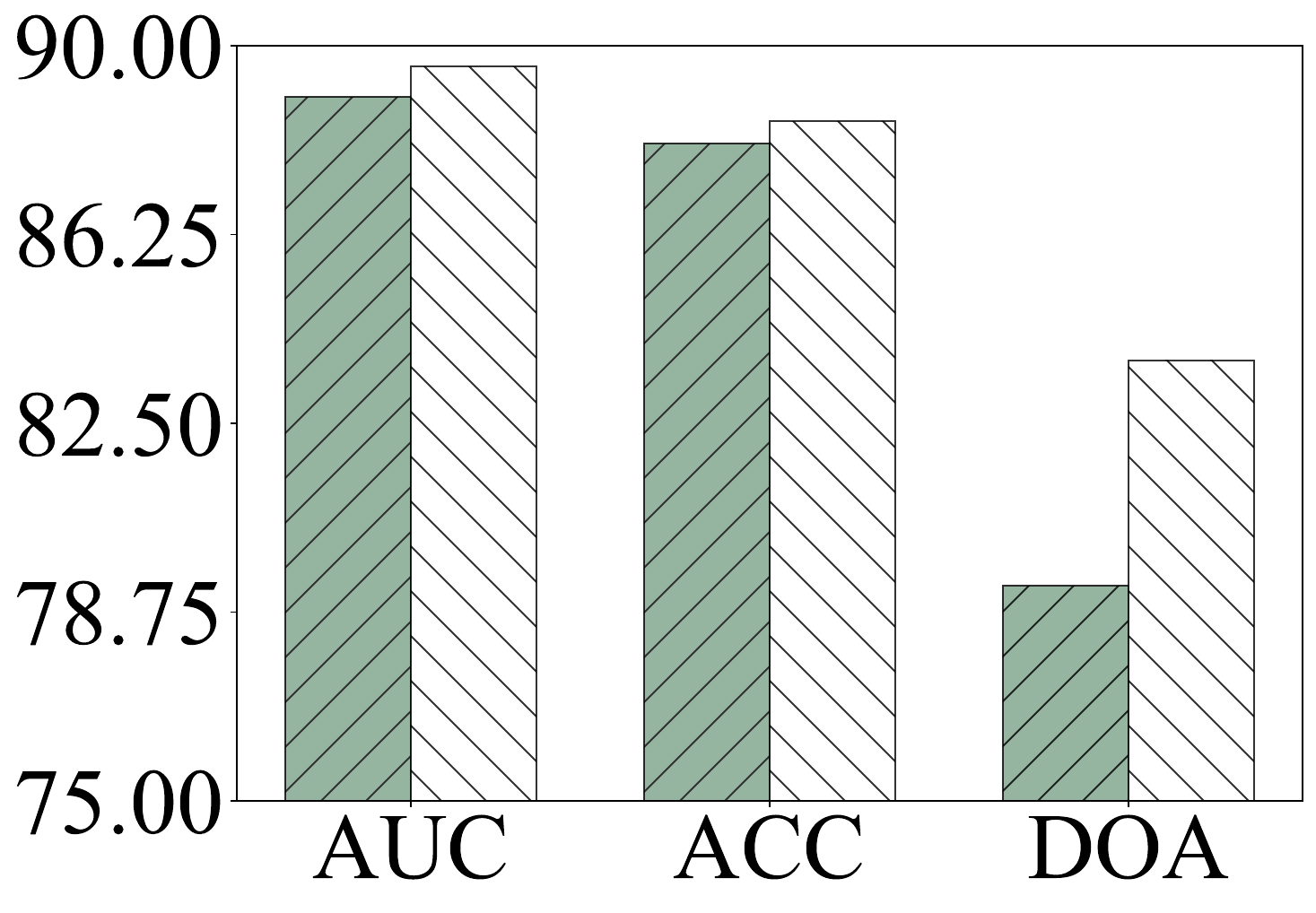}\\
    (c)  MOOC
\end{minipage}
\caption{Comparison with KSCD in the standard setting.}
\label{fig:exp:standard_kscd}
\end{figure}

In this subsection, we will compare the performance of our proposed LRCD with other baselines for Student Score Prediction in the standard setting. Specifically, the source domains and target domains originate from the same subject and platform. As shown in Figure~\ref{fig:exp:standard_or_kancd}, Figure~\ref{fig:exp:standard_kancd} and Figure~\ref{fig:exp:standard_kscd}, even in the standard setting where the source domain and target domain are identical, our method, despite not being explicitly tailored for this scenario, still demonstrates commendable performance.

\subsection{Hyperparameter Analysis}\label{appd:exp:hyper}

\textbf{The Effect of the Integrated CDMs.}
Here, we provide the results in Figure~\ref{fig:hyper_cdm}. Detailed analysis can be found in Section~\ref{sec:hyper}.

\subsection{Case Study}\label{appd:exp:case}
The student selected for the case study is from SLP-Math, identified as ``00ad006d0bcf06158f49fb0580Abd957''. We believe that student profile editing is highly beneficial in real educational scenarios. It allows students to preview their potential improvements in advance, allowing them to choose appropriate exercises to focus on, thereby reducing their academic burden. Detailed analysis can be found in the Student Profile Editing subsection in Section~\ref{sec:case}.
\end{document}